\documentclass[times,twocolumn,final]{elsarticle}

\usepackage{geometry}
\usepackage{framed,multirow}

\usepackage{amssymb}
\usepackage{latexsym}
\usepackage{amsmath}
\usepackage{ulem}
\usepackage{hhline}
\usepackage{float}
\usepackage[switch]{lineno}
\usepackage{url}
\usepackage{xcolor}

\usepackage{hyperref}

\usepackage{subcaption}

\definecolor{newcolor}{rgb}{.8,.349,.1}

\newcommand{\calB}{\ensuremath{\mathcal{B}}}
\newcommand{\calD}{\ensuremath{\mathcal{D}}}
\newcommand{\calF}{\ensuremath{\mathcal{F}}}

\newcommand{\calL}{\ensuremath{\mathcal{L}}}
\newcommand{\calM}{\ensuremath{\mathcal{M}}}
\newcommand{\calR}{\ensuremath{\mathcal{R}}}
\newcommand{\calJ}{\ensuremath{\mathcal{J}}}

\newcommand{\R}{\ensuremath{\mathbb{R}}}
\newcommand{\F}{\ensuremath{\calF}}
\newcommand{\M}{\ensuremath{\calM}}
\newcommand{\y}{\ensuremath{y}}

\definecolor{Lobe1}{RGB}{0,145,168}
\definecolor{Lobe2}{RGB}{181,22,33}
\definecolor{Lobe3}{RGB}{250,187,0}
\definecolor{Lobe4}{RGB}{236,116,4}
\definecolor{Lobe5}{RGB}{149,188,14}
\definecolor{Meancolor}{RGB}{144,99,41}

\journal{Medical Image Analysis}

\begin{document}

%\verso{Alessa Hering \textit{et~al.}}

\begin{frontmatter}

\title{CNN-based Lung CT Registration with Multiple Anatomical Constraints}%
%\tnotetext[tnote1]{This is an example for title footnote coding.}\tnoteref{tnote1}

\author[1,2]{Alessa Hering\corref{cor1}}
\ead{alessa.hering@mevis.fraunhofer.de}
\cortext[cor1]{Corresponding author:
  Tel.: +49 451 3101 6112;
  fax: +49 451 3101 6104;}
\author[1]{Stephanie Häger}
\author[3]{Jan Moltz}
\author[2]{Nikolas Lessmann}
\author[1]{Stefan Heldmann}
\author[2,3]{Bram van Ginneken}

\address[1]{Fraunhofer Institute for Digital Medicine MEVIS, Maria-Goeppert-Str 3, 23562 Lübeck, Germany}
\address[2]{Diagnostic Image Analysis Group, Radboud University Medical Center, Nijmegen, The Netherlands}
\address[3]{Fraunhofer Institute for Digital Medicine MEVIS, Am Fallturm 1, 28359  Bremen, Germany}

\begin{abstract}
%%%
Deep-learning-based registration methods emerged as a fast alternative to conventional registration methods. However, these methods often still cannot achieve the same performance as conventional registration methods because they are either limited to small deformation or they fail to handle a superposition of large and small deformations without producing implausible deformation fields with foldings inside.

In this paper, we identify important strategies of conventional registration methods for lung registration and successfully developed the deep-learning counterpart. We employ a Gaussian-pyramid-based multilevel framework that can solve the image registration optimization in a coarse-to-fine fashion. Furthermore, we prevent foldings of the deformation field and restrict the determinant of the Jacobian to physiologically meaningful values by combining a volume change penalty with a curvature regularizer in the loss function. Keypoint correspondences are integrated to focus on the alignment of smaller structures.

We perform an extensive evaluation to assess the accuracy, the robustness, the plausibility of the estimated deformation fields, and the transferability of our registration approach. We show that it achieves state-of-the-art results on the COPDGene dataset compared to conventional registration method with much shorter execution time. In our experiments on the DIRLab exhale to inhale lung registration, we demonstrate substantial improvements (TRE below $1.2$ mm) over other deep learning methods. Our algorithm is publicly available at https://grand-challenge.org/algorithms/deep-learning-based-ct-lung-registration/. 
%%%%
\end{abstract}


\end{frontmatter}

%\linenumbers

%% main text
\section{Introduction}
\label{sec:Introction}
Image registration is the process of aligning two or more images to achieve point-wise spatial correspondence. This is a fundamental step for many medical image analysis tasks and has been an active field of research for decades \citep{maintz1998surveyIR,SotirasDavatzikosParagios2013}. Various approaches and tailored solutions have been proposed to a wide range of problems and applications. Typically, image registration is phrased as an optimization problem with respect to a spatial mapping that minimizes a suitable cost function and common approaches estimate solutions by applying iterative optimization schemes. Unfortunately, solving such an optimization problem is computationally demanding and consequently slow.

While deep learning has become the methodology of choice in many areas, relatively few deep-learning-based image registration algorithms have been proposed. One reason for this is the lack of ground truth and the large variability of plausible deformations that can align corresponding anatomies. Therefore, the problem is much less supervised than for example image classification or segmentation. Nevertheless, several methods have been presented in the last years which aim to mimic the process of conventional image registration methods by training a neural network to predict the non-linear deformation function given two new unseen images.
As a trained neural networks can process images in real time, this has immense potential for time-sensitive applications such as image guidance in radiotherapy, tracking, or shape analysis through multi-atlas registration. \\

In this paper, we target the challenging task of lung registration. The complexity of this registration task is manifold, as the occurring motion is a superposition of respiratory and cardiac motion.
Moreover, the sliding motion between the lung and rib cage during breathing -- more precisely between pleura visceralis and pleura parietalis -- is an additional challenge. The scale of the motion within the lungs can often be larger than the structures (vessels and airways) that are used to guide the optimization process. This may cause a registration algorithm to get trapped in a local minimum \citep{heinrich2013mrf,polzin2013combining}. This makes the problem even more difficult.Therefore, a registration method needs to be able to estimate a displacement field that accounts for substantial breathing motion but also aligns small structures like individual pulmonary blood vessels precisely.

\section{Related Work}
Most deep-learning-based approaches aim to learn a registration function in form of a convolutional neural network to predict spatial deformations warping a moving image to a fixed image. All these works have contributed improving deep-learning-based image registration and have been applied to different registration applications including brain MR \citep{balakrishnan2018unsupervised,balakrishnan2019voxelmorph,yang2017quicksilver}, cardiac MR \citep{deVos2017DIRnet}, cardiac MR-CT \citep{hering2019memory}, prostate MR-US \citep{hu2018labeldriven}, thorax-abdomen CT \citep{heinrich2019closing}, thorax CT \citep{deVos2019deep,eppenhof2018pulmonary, eppenhof2019progressively,hering2019SPIE,hering2019mlvirnet,sentker2018gdl} and CT-CBCT registration \citep{kuckertz2020ctcbct}. Existing approaches can be classified as \textit{supervised}, \textit{unsupervised}, and \textit{weakly-supervised} techniques based on how much supervision is available.\\

\textit{Supervised methods} use ground-truth deformation fields for training. The ground truth can be generated in different ways. In \cite{eppenhof2018deformable} and \cite{sokooti2017nonrigid} the network is trained on synthetic random transformations. A drawback is that the randomly generated ground truth is artificial and may not be able to reproduce all possible deformations. Alternatively, conventional registration methods can be used to produce deformations by registering images \citep{sentker2018gdl,YangEtAl2017} or other image features like landmarks or segmentations \citep{rohe2017svf}. Another way to create a ground truth is to combine simulations with existing algorithms \citep{krebs2017supervised}. Consequently, the performances of all these approaches is upper bounded by the quality of the initial registration algorithm or the realism of the synthetic deformations.

In contrast, \textit{unsupervised methods} \--- also called \textit{self-supervised methods} \--- do not require any ground truth. The idea is to use the cost function of conventional image registration (similarity measure and regularization term) as the loss function to train the neural network. An important milestone for the development of these methods was the introduction of the spatial transformer network \citep{jaderberg2015spatial} to differentiably  warp images. This differentiable warping has actually been part of most conventional registration methods for a long time (e.g. \cite{koenig2018matrixfree,modersitzki2004numerical,modersitzki2009fair}). The concept of an unsupervised deep-learning-based registration method was first introduced with the DIRNet \citep{deVos2017DIRnet} for 2D image registration using the normalized cross-correlation image similarity measure as loss function. In \cite{li2018nonrigid} the approach has been extended by adding diffusion regularization to the loss function forcing smooth deformations. The method has successfully been demonstrated for registration of 3D brain subvolumes. The idea of unsupervised deep-learning-based image registration has been further evolved in several works \citep{balakrishnan2018unsupervised,deVos2019deep,ferrante2018adaptability,hering2019SPIE,krebs2019unsupervised}.

\textit{Weakly-supervised methods} do not rely on ground-truth deformation fields either but training is still supervised with prior information. In \cite{hu2018labeldriven} and \cite{hu2018weakly}, a set of anatomical labels is used in the loss function. The labels of the moving image are warped by the deformation field and compared with the fixed labels. All anatomical labels are only required during training. In \cite{balakrishnan2019voxelmorph} and \cite{hering2019mlvirnet,hering2019memory}, the complementary strengths of global \mbox{semantic} information and local distance metrics were combined to improve the registration accuracy. %In \cite{heinrich2019closing}, a ....

In conventional registration approaches, multilevel continuation and scale-space techniques have been proven very efficient to avoid local minima during the optimization process of the cost function, to reduce topological changes or foldings, and to speed up runtimes \citep{bajcsy1989multiresolution,haber2004cofir,kabus2010multilevel,schnabel2001generic} -- explaining the popularity of multi-level strategies in conventional registration methods. 
As a lot of deep-learning-based registration methods are build on top of U-Net (e.g. \cite{balakrishnan2019voxelmorph,hering2019SPIE,Hering2019BVM,rohe2017svf}), they are also multi-leveled in their nature. 
%. For that reason, most conventional registration approaches are using a multilevel strategy. The first deep-learning-based registration approaches make use of a multilevel strategy as they are based on the U-Net architecture \citep{balakrishnan2019voxelmorph,hering2019SPIE,Hering2019BVM,rohe2017svf}. 
The first half of the "U" (the encoder) generates features on different scales starting at the highest resolution and reducing the resolution through pooling operations. In this procedure, however, only feature maps on different levels are calculated but neither are different image resolutions used nor deformation fields computed. Only a few approaches implement a multi-resolution or hierarchical strategy in the sense of multilevel strategies associated with conventional methods. In \cite{hu2018labeldriven}, the authors proposed an architecture that is divided into a global and a local network, which are optimized together. In \cite{eppenhof2019progressively}, a multilevel strategy is incorporated into the training of a  U-Net, by growing and training progressively level-by-level. In \cite{deVos2019deep}, a patch-based approach is presented, where multiple  CNNs (ConvNets) are combined additively into a larger architecture for performing coarse-to-fine image registration of patches. The results from the patches are then combined into a deformation field warping the whole image. Another patch-based multilevel approach is presented in \cite{fu2020lungregnet}. The multilevel framework consists of a CoarseNet and a FineNet which are trained jointly. During training, the estimated deformation field of the CoarseNet and the FineNet are not combined but the moving patch is transformed twice. During inference, if the mean absolute differences between the deformed image patch and the fixed image exceeds a predefined threshold, FineNet is applied again. This leads to a variable number of deformation field patches, which are combined additively. Although previous deep-learning-based registration works (e.g. \cite{deVos2019deep,eppenhof2019progressively,sentker2018gdl}) contributes many efforts to improve the registration accuracy for lung registration, there is still a misalignment of smaller structures in the lung, which leads to a high target registration error of landmarks. 

%In \cite{jiang2020multi}, a similar strategy is used, however, instead of training each CNN progressively, all CNN models are jointly trained by composing the loss of all levels. The deformation fields are additively combined like in \cite{deVos2019deep}. 

\section*{Contribution}
We previously introduced an end-to-end deep-learning multilevel registration method that can handle large deformations by computing deformation fields on different scales and functionally composing them \cite{hering2019mlvirnet}. This initial study, despite its limited evaluation, proved that it is a valid strategy to improve the alignment of vessels and airways -- though a gap regarding the target registration error of landmarks with the best conventional registration methods remained. Building on this previous work, and addressing its limitations, we were able to further close that gap.
 
Our key contributions are as follow:
\begin{itemize}
    \item We present multiple anatomical constraints to incorporate anatomical priors into the registration framework to obtain more realistic results. We integrate the lung lobe mask to consider the global context. Moreover, the keypoint correspondences are used to increase the alignment of airways and vessels.
    \item We introduce a novel constraining method to control volume change and therefore avoid foldings inside the deformation field. While the idea of volume change control is not new in conventional registration, we firstly present a suitable version for deep-learning-based image registration.
    \item We perform comprehensive experiments on three different datasets -- the multi-center COPDGene study \citep{regan2011COPDGene} and the DIRLab challenge dataset \citep{castillo2009DIRLAB,castillo2013DIRLAB}, and the EMPIRE10 challenge dataset \citep{murphy2011evaluation} -- to assess the accuracy, plausibility, robustness, transferability of our method. We achieve  comparable results as state-of-the-art registration approaches.
\end{itemize}

\section{Method}
\begin{figure*}
\centering
\includegraphics[width=\textwidth]{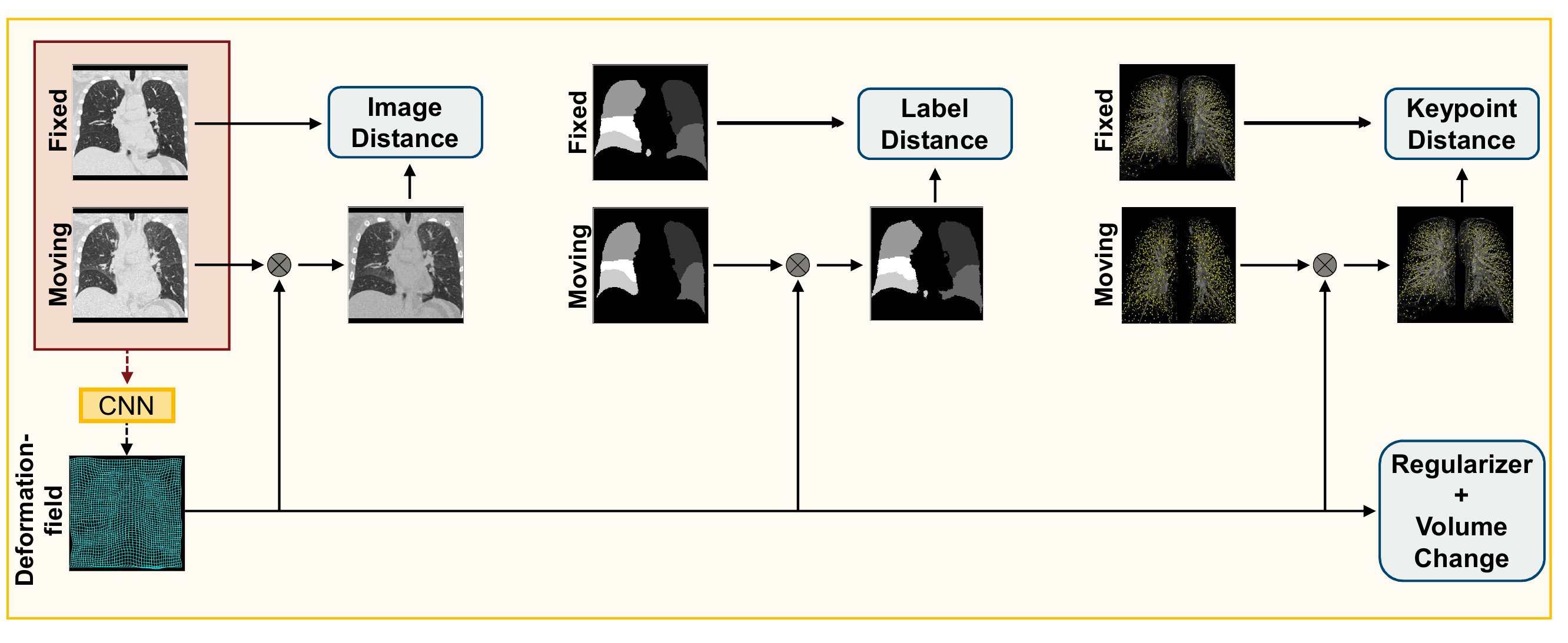}
\caption{Schematic representation of the training process. In the loss function, we compare the fixed image, pulmonary lobes mask and keypoints to the deformed moving image, pulmonary lobes mask and keypoints, respectively. To enforce smoothness and to prevent foldings, a regularizer and a volume change penalty are integrated into the loss function. During inference, only the fixed and moving image is required to estimate the deformation field. For a better visualization, we have placed the windowed CT image in the background of the used keypoints. Best viewed in colors.}
\label{fig:scheme}

\end{figure*}
\subsection{Variational Registration Approach}
Let $\F,\M:\R^3\to\R$ denote the fixed image and moving image, respectively, and let $\Omega\subset\R^3$ be a domain modeling the field of view of $\calF$. Registration methods aim to compute a deformation $\y:~\Omega\to~\R^3$ that aligns the fixed image $\calF$ and the moving image $\calM$ on the field of view $\Omega$ such that $\calF(x)$ and $\calM(\y(x))$ are similar for $x\in\Omega$. The deformation is defined as a minimizer of a suitable cost function that typically takes the form
\begin{equation}\label{eq:Jfun}
\calJ(\calF,\calM,\y)~=~\calD(\calF,\calM(\y))~+~\alpha\calR(\y)
\end{equation}
with so-called distance measure $\calD$ that quantifies the similarity of fixed image $\calF$ and deformed moving image $\calM(y)$ and so-called regularizer $\calR$ that forces smoothness of the deformation typically by penalizing spatial derivatives. Typical examples for the distance measure are the squared $L_2$ norm of the difference image (SSD), normalized cross correlation (NCC), or mutual information (MI). The cost function can be extended by additional penalty terms to force desired properties or incorporate additional knowledge in form of anatomical constraints \citep{ruhaak2017estimation}. As illustrated in Figure \ref{fig:scheme}, our method inputs both the fixed and moving image into the network that predicts the dense displacement field. 
The loss function uses all available information: input images, segmentation masks and keypoints, with additional regularization -- in the form of a smoothness prior and a volume consistency constraint -- to prevent foldings.

\subsection{Loss Function}

\subsubsection{Normalized Gradient Field Distance Measure}
One of the main challenges of lung registration are the varying intensity changes occurring due to the altered density of lung tissue during breathing. This leads to a violation of the intensity constancy assumption between corresponding points, on which the classic sum of squared differences (SSD) distance measure is built. However, the lung exhibits a rich structure of bronchi, fissures, and especially vessels that can be exploited for the registration, more suited to distance measure that focus on image edges rather than intensities. We follow the approach of \cite{ruhaak2017estimation} and \cite{hering2019mlvirnet} using the \textit{normalized gradient fields} (NGF) \citep{HaberModersitzki2006} distance measure

\begin{equation*}
    \mathcal{D}(\F,\M(\text{y})) = \int_\Omega 1- \frac{\langle \nabla \M(\text{y}(x)), \nabla \F(x)\rangle^2_\epsilon}{\Vert\nabla\M(\text{y}(x))\Vert^2_\epsilon \Vert\nabla\F(x)\Vert^2_\epsilon} \, \text{d}x,
\end{equation*}
with $\langle f,g \rangle_\epsilon := \sum_{j=1}^3 (f_j g_j +\epsilon^2)$,
$\|f\|_\epsilon := \sqrt{\langle f, f \rangle_\epsilon}$. 
The edge hyper-parameter $\epsilon > 0$ is used to suppress small image noise, without affecting image edges. Therefore, a good strategy is to choose its value relative to the average gradient. In \citep{HaberModersitzki2006}, the following automatic choice is suggested:
\begin{equation*}
    \epsilon = \frac{\nu}{V} \int_\Omega \| \nabla I(x)\| dx,
\end{equation*}
where $\nu$ is the estimated noise level in the image and $V$ is the volume of the domain $ \Omega$. For CT images, a value in the range of $\left[0.1, 10\right]$ is mostly a good choice.

Since we focus on accurate registration inside the lungs and to avoid misalignment artifacts due to sliding motion at the pleura, we restrict $\Omega$ to the support of the lung mask of the fixed image. \\

\subsubsection{Curvature Regularizer}
Smooth deformation fields are enforced by the second order \textit{curvature regularizer}~\citep{fischer2003curvature} given by
\begin{equation}\label{eq:curv}
    \mathcal{R}(\text{y}) =  \int_\Omega \sum_{j=1}^3 \Vert \Delta \text{y}_j(x) \Vert^2 \, \text{d}x.
\end{equation}

\subsubsection{Volume Change Control}
\begin{figure*}[h!]
        \centering
        \begin{subfigure}[b]{0.33\textwidth}
                \includegraphics[width=\textwidth]{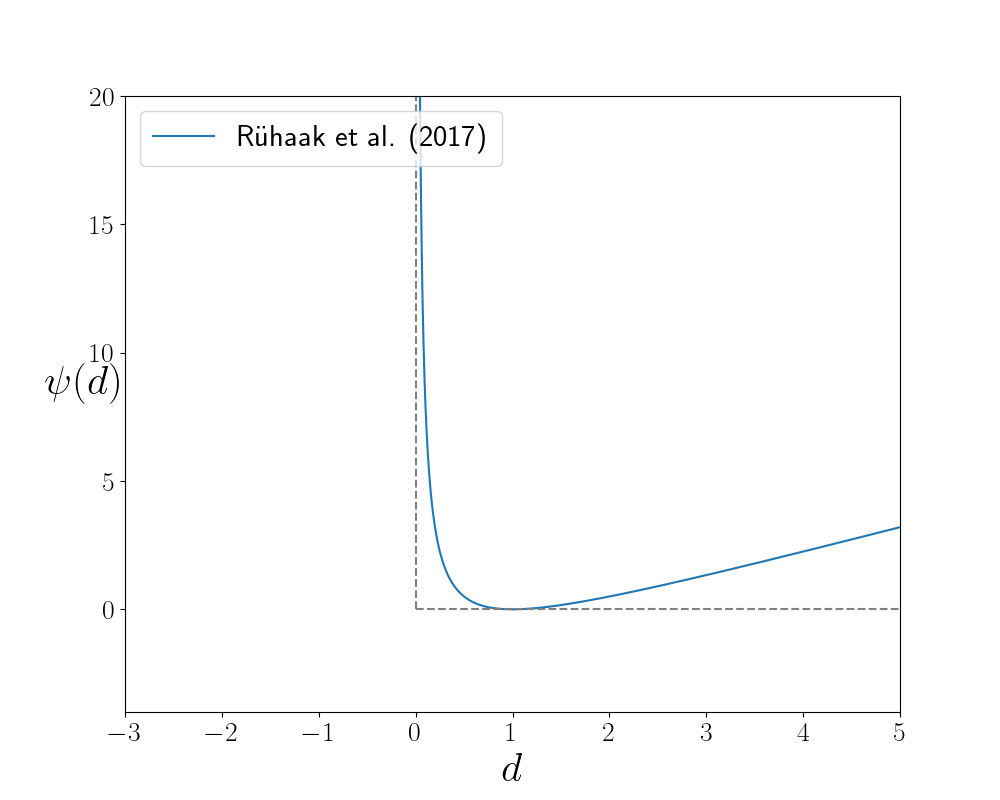}
                \caption{Interior point method from \cite{ruhaak2017estimation}}
        \end{subfigure}
        \begin{subfigure}[b]{0.33\textwidth}
                \includegraphics[width=\textwidth]{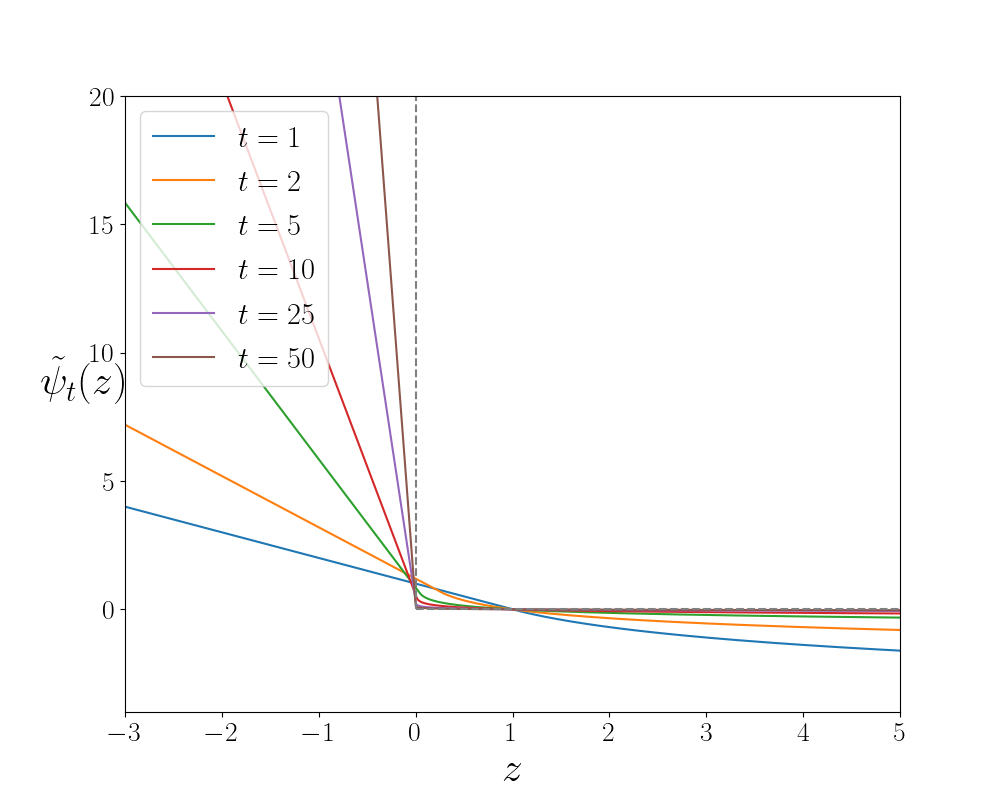}
                \caption{Log-barrier extension from \cite{kervadec2020bounding}}
        \end{subfigure}
        \begin{subfigure}[b]{0.33\textwidth}
                \includegraphics[width=\textwidth]{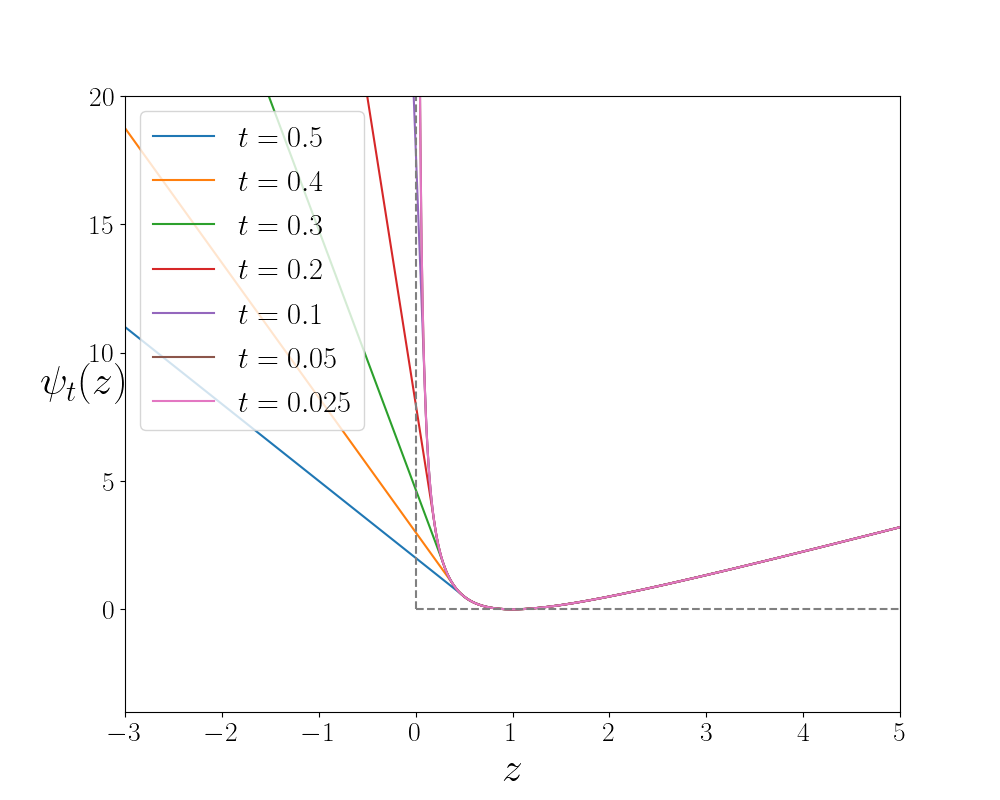}
                \caption{Our proposed penalty}
        \end{subfigure}
        \caption{A graphical illustration of both standard log-barrier (a), the proposed log-barrier extension (b) and examples of penalty functions (c). The solid curves in colors show the approximations for several $t$ values of functions $\tilde \psi_t (z)$ and $\psi_t (z)$  respectively. }
        \label{fig:logBarrier}
\end{figure*}

Although the curvature regularization from Equation \eqref{eq:curv} prefers smooth deformation, foldings may still happen, which is obviously physically impossible. More formally, foldings happen when the Jacobian determinant of the deformation field becomes negative. 
To avoid any foldings, we therefore aim to minimize the distance measure $\mathcal D$ and the regularizer $\mathcal R$ while keeping the Jacobian determinant positive, for every voxel in $\Omega$. Formally, this can be written as a constrained optimization problem:
\begin{alignat*}{2}
   \min_{\y} \quad & \calD(\calF,\calM(\y))~+~\alpha\calR(\y) \\
  \text{s.t.} \quad &  \begin{aligned}[t]
     \text{det}\nabla y(x) & > 0 \quad \ \quad \forall x \in \Omega.
  \end{aligned}
\end{alignat*}
To achieve this, \cite{ruhaak2017estimation} introduced a \textit{Volume Change Control} (VCC) that could be integrated in their overall objective:
\begin{equation}\label{eq:VCC}
    \mathcal{V}(y) = \int_{\Omega} \psi(\text{det}\nabla y(x)) \text{d}x,
\end{equation}
where
\begin{equation}\label{eq:ruhaak}
    \psi(z) =
    \begin{cases}
        \frac{(z-1)^2}{z} & \text{if} \ z > 0 \\
        +\infty & \, \text{otherwise.}
    \end{cases}
\end{equation}
For the sake of simplicity, the input to $\psi$ in equation \ref{eq:VCC} is substituted with $z=\text{det}\nabla y(x)$ in equation \ref{eq:ruhaak}.
Notice that $\psi(z)$ is minimized when $z=1$ (see Figure \ref{fig:logBarrier} (a)). Therefore, the regularizing effects of the VCC are twofold: i) prevents the formation of foldings, by keeping the determinants positive, ii) limits both shrinkage and expansions by biasing the optimization to keep the same volume.
%\textcolor{red}{Between us, notice that another way to write $\psi(d)$ is $\psi(d) = +\infty_{[d \leq 0]} + 1_{[d > 0]}\frac{(d-1)^2}{d}$;which I find highlights more the two-fold objectives of that one.}

The method that \cite{ruhaak2017estimation} used falls into the category of \textit{interior-point methods}. Such methods became very popular in constrained optimization \citep{boyd2004convex} as they do not require the expansive primal-dual updates of traditional Lagrangian optimization: the infinity penalty acts as a "barrier", preventing the optimization to go out of bounds.

% Despite their many successes and optimality guarantees, 
To be used, interior-points methods require a feasible starting point: all constraints need to be strictly satisfied before starting the optimization procedure. This is usually done in a pre-optimization step (called Phase I) before the actual optimization of Phase II is performed. We can see it as finding a valid initial guess, and then refining it. 

In the context of deep neural networks, standard Lagrangian methods are not feasible due to their expensive primal-dual updates, which requires to retrain a neural network (from scratch) at each iteration. Interior-point methods are also not applicable, as solving phase I requires to solve a constrained optimization problem in the first place. 

\cite{kervadec2019constrained} proposed a parametric log-barrier \textit{extension} (illustrated in Figure \ref{fig:logBarrier} (b)), that does not require an initial feasible solution:
\begin{equation}\label{eq:logBarrier}
    \tilde{\psi}_t(z) =
    \begin{cases}
-\frac{1}{t} \log (z) & \text{if} \ z \geq \frac{1}{t^2} \\
-tz -\frac{1}{t}\log\left(\frac{1}{t^2}\right) + \frac{1}{t} & \, \text{otherwise.}
\end{cases}
\end{equation}
$t$ is a hyper-parameter, controlling the slope of the barrier. By starting with a small initial value, and increasing it as the training progresses, one is able to "raise" the barrier, closing it eventually. 

We propose to keep the property of equation \eqref{eq:ruhaak} to symmetrically penalizes local shrinkage and expansion and make it applicable for neural networks by using the barrier formulation of equation \eqref{eq:logBarrier} for $z<t$ with $t\rightarrow 0 $ over time (illustrated in Figure \ref{fig:logBarrier} (c)): 

\begin{equation}\label{eq:volgBarrier}
    \psi_t(z) =
    \begin{cases}
 \frac{(z-1)^2}{z} &  \text{if} \ z \geq t \\
\left(1-\frac{1}{t^2}\right)z+\frac{2(1-t)}{t} & \, \text{otherwise,}
\end{cases}
\end{equation}
with $t>0$ which is a hyper-parameter controlling the slope of the linear barrier for $z<t$. This barrier can be raised during the training by decreasing the value of $t$ to penalize foldings more strongly. Note that the linear part for $z<t$ is chosen such that $\psi$ is continuously differentiable provided $t>0$. In our experiments, we set $t=0.2$ for the first level of our multilevel architecture and decrease it by the factor of 2 for any further level. For $z~\geq~t$, we symmetrically penalize local shrinkage and expansion, i.e., $\psi(z)=\psi(1/z)$.

%\textcolor{red}{What I don't get is why not simply minimize $\tilde\psi_t(\frac{(d-1)^2}{d})$, or even $\tilde\psi_t(d) + max(0, \frac{(d-1)^2}{d})$}

\subsubsection{Mask Alignment}
Several recent publications (e.g. \cite{balakrishnan2019voxelmorph,hering2019memory}) have shown that adding further information in the form of segmentation masks into the loss function can guide the network during the training process. Since the segmentation masks are used in the loss function, they are only required during training and not for registration of unseen images. We integrate  segmentation masks by using the SSD loss
\begin{equation}\label{eq:BoundaryLoss}
    \calB(\y) = \frac{1}{2} \int_{\Omega} \|b_\M(\y(x))-b_\F(x)\|^2 \text{d}x,
\end{equation}
%\begin{equation*}
%   \calB(\y) =  1- %\frac{2b_\F(x)*b_\M(y(x))}{\|b_\F(x)\|\|b_\M(y(x))\|}
%\end{equation*}\label{eq:BoundaryLoss}

where $b_\F :\Omega\rightarrow\left[0,1\right]^k$ and $b_\M :\Omega\rightarrow\left[0,1\right]^k$ denote functions of $\F$ and $\M$ that are the one-hot representation of the segmentation mask, with $k$ the number of different labels. For lung registration, we use segmentation of the lungs into the five pulmonary lobes ($k=5$). During training, we use linear interpolation to warp the one-hot segmentation masks since this results in a smoother loss function at the border of the segmentation. With nearest neighbor interpolation, the loss of each voxel can either be one or zero. Linear interpolation allows for probabilistic loss values between zero and one.

\subsubsection{Keypoint Loss}
For conventional image registration, previous work (e.g. \cite{ehrhardt2010automatic,polzin2013combining,ruhaak2017estimation}) has shown that the integration of sparse keypoints during the optimization of the deformation field yields better registration results. In contrast to conventional registration approaches, keypoints can be integrated into the loss function and are therefore, similar to the segmentation masks for the mask penalty, only needed for training but not during inference. In general, there are several ways to integrate the keypoints into an intensity-based registration approach (e.g. \cite{fischer2003combination}, \cite{ruhaak2017estimation}, \cite{papenberg2009landmark}). We choose to integrate the keypoint information through a least-squares penalty into our model by directly comparing the transformed keypoint of the fixed image with the corresponding moving keypoint:
\begin{equation*}
    \mathcal{K}(\y) = \frac{1}{|K|}\sum_{i=1}^{|K|} \left\| k_\mathcal{M}^i-\y\left(k_\mathcal{F}^i\right)   \right\| ^2
\end{equation*}
with the moving keypoint $k_\mathcal{M}^i$ and the warped fixed keypoint $y\left(k_\mathcal{F}^i\right)$ for all $|K|$ keypoints. In general, manually annotated landmarks or automatically generated keypoints can be integrated with this loss function. However, since manual annotation of landmarks is time-consuming, we use the keypoint detection algorithm described in \cite{ruhaak2017estimation} to generate a large number of corresponding keypoints.\\

The final loss is given by
\begin{equation}\label{eq:Loss}
\calL(\calF,\calM,\y)~=~\calD(\calF,\calM(\y))~+~\alpha\calR(\y)~+~\beta\calB(\y)+\gamma\mathcal{V}(\y)~+~\delta\mathcal{K}(\y).
\end{equation}
The hyper-parameters $\alpha, \beta, \gamma$ and $\delta$ have to be chosen manually. However, our experiments showed that a change in the magnitude leads to only slight changes in the results.

\subsection{Baseline Architecture}
Our CNN is based on a U-Net \citep{ronneberger2015unet} which takes the concatenated 3D moving and fixed image as input and predicts a 3D dense displacement field with the same resolution as the input images. The U-net consists of three levels starting with 16 filters in the first layer, which are doubled after each downsampling step. We apply 3D convolutions in both encoder and decoder path with a kernel size of 3 followed by an instance normalization and a ReLU layer. In the encoder path, the feature map downsampling steps use $2~\times~2~\times~2$ average pooling with a stride of 2. In the decoder path, the upsampling steps use transposed convolution with $2~\times~2~\times~2$ filters and half the number of filters than the previous step. The final layer uses a 1x1x1 convolution filter to map each 16-component feature vector to a three-dimensional displacement.

\subsection{Multilevel Architecture}
In conventional image registration, multilevel continuation has been proven very efficient to avoid local minima, to reduce topological changes or foldings, and to speed up runtimes \citep{bajcsy1989multiresolution,haber2004cofir,kabus2010multilevel,schnabel2001generic}. Recent deep-learning-based approaches \citep{deVos2019deep,fu2020lungregnet,hering2019mlvirnet,jiang2020multi,mok2020large} have shown that, besides carrying over these properties, a multilevel scheme helps overcome the limitations of deep-learning-based registration approaches to properly deal with small and local deformations.

\begin{figure*}
\centering
\includegraphics[width=0.65\textwidth]{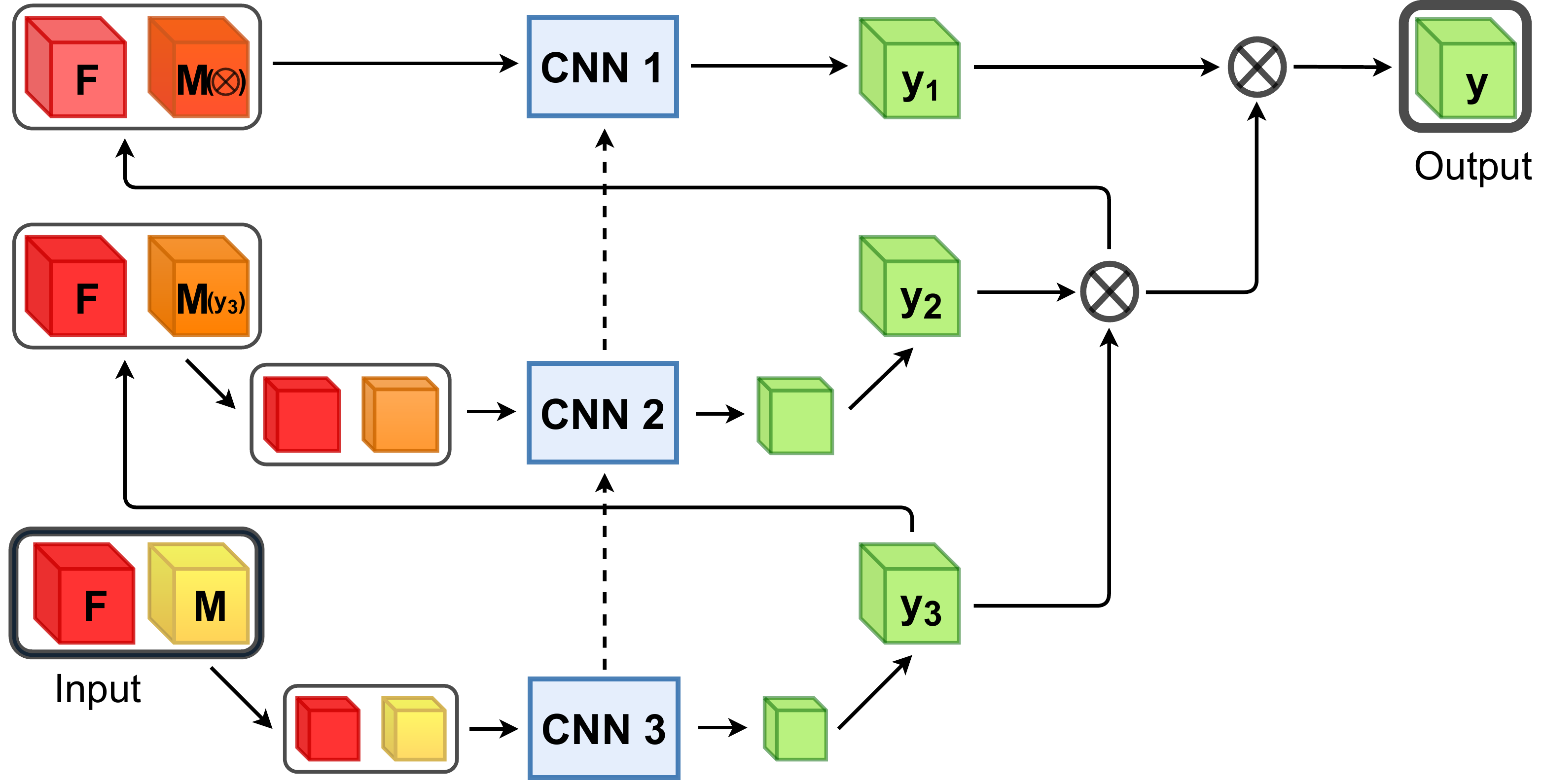}
\caption{Overall scheme of the proposed multilevel framework, where $\F$ indicates the fixed image, $\M$ the moving image, $\y$ the deformation field and $\M (y)$ the warped image. Each CNNs is trained separately for a fixed amount of epochs and the weights stay fixed afterwards. The deformation fields from all preceding coarse levels are used as an initial guess by combining them by functional composition and warp the moving image on the highest resolution. Subsequently, the warped moving image is downsampled to the current image level. The dotted lines illustrate the initialization of the network weights with the learned parameters of the previous level. }
\label{fig:MultilevelFramework}
\end{figure*}

As in our previous work \citep{hering2019mlvirnet}, we follow the ideas of standard multilevel registration and compute coarse grid solutions that are prolongated and refined on the subsequent finer level.
Our multilevel framework is illustrated in figure \ref{fig:MultilevelFramework} with $L=3$ levels. The registration starts on the coarsest level $L$ where the deformation $\tilde{y}_L$ is computed from the input images that have been Gaussian-smoothed and downsampled by a factor of $2^{L-1}$. On all finer levels $\ell<L$, we incorporate the deformations from all preceding coarse levels as an initial guess by combining them by functional composition and warping the moving image. Subsequently, the fixed and warped moving images are downsampled. \\

The number of used levels is a hyper parameter which should be chosen depending on the task and the used data. The maximal number of levels that can be used is limited by the GPU memory and the image size. Since the images are downsampled with a factor of two in the multilevel setting and additionally the image features are downsampled three-times in the U-Net, the number may be chosen at most so that the image size is divisible by $2^{3+(L-1)}$. Our experiments (c.f. section \ref{sec:AblationStudy}) have shown that a three-level scheme works best in our application and fits on a 12GB GPU. In our experiments, we use in particular a three-level scheme ($L=3$, Fig. \ref{fig:scheme}). The three networks are trained progressively. First, the network on the coarsest level is trained for a fixed amount of epochs. Afterwards, the parameters of the middle network are learned while the coarsest network stays fixed and is only used to produce the initial deformation field. The same procedure is repeated on the finest level. The same architecture is used on all levels. The network parameters on the coarsest level are initialized with Xavier uniform \citep{xavier2010initialization}, whereas all other networks are initialized with the learned parameters of the previous network. Note that the receptive field in voxels is the same for all networks, however, due to the decreased resolution on the coarse levels, the receptive field in mm is much larger.

\section{Experiments}
We perform several experiments to assess the accuracy, plausibility, robustness, transferability, and speed of our weakly-supervised deep-learning-based registration approach.

\subsection{Data}
We train and validate our method on the data from the COPDGene study \citep{regan2011COPDGene}. To prove the robustness and transferability of our method and to compare our method with other registration approaches, we evaluate our registration approach on the publicly available DIRLab dataset~\citep{castillo2009DIRLAB,castillo2013DIRLAB} and on the EMPIRE10 challenge as well. On the COPDGene dataset, the evaluation is based on the lobe segmentation masks, and on both of the other datasets, annotated landmarks are available on which we evaluate the target registration error. \\

\subsubsection{COPDGene Dataset}
\begin{figure}[h]
\centering
\setlength{\tabcolsep}{0.001\textwidth}
\begin{tabular}{cc}
  \includegraphics[width=0.22\textwidth]{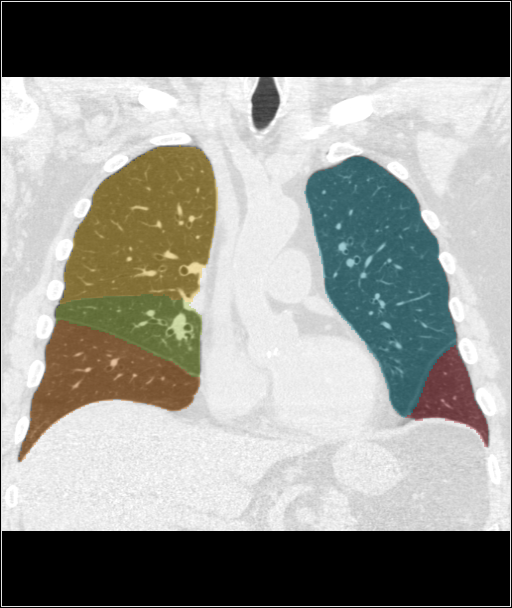}
& \includegraphics[width=0.22\textwidth]{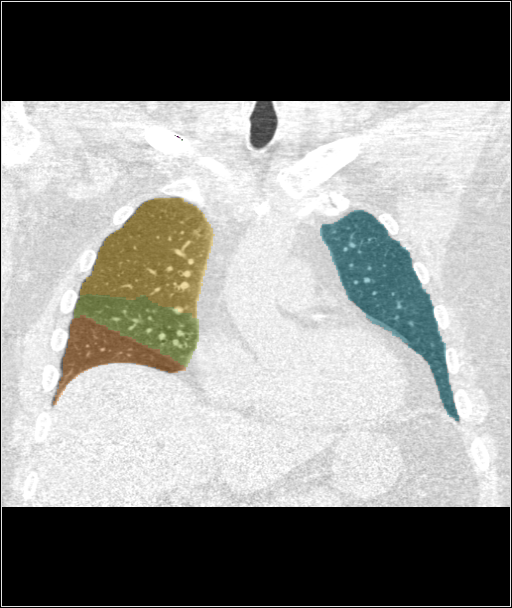}
\\
  {a) Inspiration}
& {b) Expiration}
\end{tabular}
\caption{The Image shows a) an inspiration scan and b) and expiration scan of the lungs subdivided into \textcolor{Lobe1}{\rule{.2cm}{.2cm}} upper left lobe, \textcolor{Lobe2}{\rule{.2cm}{.2cm}} lower left lobe, \textcolor{Lobe3}{\rule{.2cm}{.2cm}} upper right lobe, \textcolor{Lobe4}{\rule{.2cm}{.2cm}} lower right lobe and \textcolor{Lobe5}{\rule{.2cm}{.2cm}} middle right lobe.} \label{fig:lobes}
\end{figure}

Training, validation, and testing data were acquired from the COPDGene study, a large multi-center clinical trial with over 10,000 subjects with chronic obstructive pulmonary disease (COPD) \citep{regan2011COPDGene}. The COPDGene study includes clinical information, blood samples, and chest CT scans. The image dataset was acquired across 21 imaging centers using a variety of scanner makes and models. Each patient had received two breath-hold 3D CT scans, one on full inspiration (200mAs) and one at the end of normal expiration (50mAs). About five years later, follow-up images were acquired from about 6000 subjects. In our study, we use the inspiration and expiration scans of 1000 patients. We split these patients into 750, 50, 200 patients for training, validation, and testing, respectively. The original images have sizes in the range of $512\times 512 \times \{ 341,\hdots, 974\}$ voxels.
The in-plane resolution of the axial slices varied between $0.5$mm to $0.97$mm per voxel with a slice thickness of $0.45$mm to $0.7$mm. \\
The human lungs are sub dived into five lobes that are separated by visceral pleura called pulmonary fissure. An exemplary inspiration scan and expiration scan of one patient with the lobe segmentation overlay is shown in Fig. \ref{fig:lobes}. For all scans segmentations of the lobes are available, which were computed automatically and manually corrected and verified by trained human analysts. \\

\subsubsection{DIRLab Challenge}
This dataset consists of ten thoracic 4D CT images acquired as part of the radiotherapy planning process for the treatment of thoracic malignancies. In our study we are only using the inspiration and expiration phase of the 4D image, i.e., two of the ten images per 4D scan. The in-plane resolution of the $512\times 512$ axial slices varied between $0.97$mm to $1.16$mm per voxel with a slice thickness of $2.5$mm. Each scan pair contains 300 manually annotated corresponding landmarks in the lung on which we evaluate the target registration error.

\subsubsection{EMPIRE10 Challenge}
The EMPIRE10 challenge \citep{murphy2011evaluation} consists of 30 scan pairs from six different categories: breathhold inspiration scan pairs, breathhold inspiration and expiration scan pairs, 4D data scan pairs, ovine data scan pairs, contrast-noncontrast scan pairs and artificially warped scan pairs. Further information on each category can be found in the challenge paper \citep{murphy2011evaluation}. Each scan pair contains 100 annotated corresponding landmarks.

\subsection{Preprocessing}
In this work, we focus on non-rigid, non-linear deformations and for that reason we perform a linear prealignment of fixed and moving image as preprocessing. For all methods, the same preprocessing is used. We subsequently warp and resample the moving image on the field of view and resolution of the fixed image, which yields a pre-registered moving image $\hat{\M}$. Lung regions are automatically cropped for each CT and resized to volumes of dimension $192\times 160\times 192$ as the network input. However, although the deformation field is computed from low-resolution input, during inference, the output deformation field is up-sampled to the original image resolution using trilinear interpolation and the overall evaluation is performed at full resolution of the original images. We do not perform any further preprocessing like normalization on the images, because the CT images are already in a standardized range (Hounsfield units). On the training data, we use the keypoint detection algorithm described in \cite{ruhaak2017estimation} to automatically compute keypoints inside the lung. These keypoints can be considered noisy labels with residual errors of 1-2mm

\subsection{Implementation Details}
We implemente our method in PyTorch. Each network was trained for 25 epochs on an NVIDIA Titan Xp using an ADAM optimizer with a learning rate of $10^{-3}$. The training of all three networks takes about 20 hours. We empirically chose the loss weighting parameters $\alpha=10$, $\beta=1$, $\gamma=0.01$. For the coarsest level, the keypoint weighting parameter $\delta$ was set to zero such that the network can focus on the coarse alignment of larger structures. In the subsequent levels, we chose $\delta=10^{7}$. For the edge parameter of the NGF distance measure, we chose $\epsilon=1$.

\subsection{Accuracy}
We evaluate our method by using the propagated lobe segmentation and the fixed lobe segmentation. If a deformation field represents accurate correspondences, the lobe segmentation of the fixed image $b_\F$ and the warped lobe segmentation of the moving image $b_\M(y)$ should overlap well. In contrast to a lung segmentation overlap, the lobe segmentation overlap provides information about inner lung structures. A good alignment of the lobes was shown to be indicative of good alignment of the fissures, which the evaluation of registration quality in \cite{murphy2011evaluation} has shown to be indicative of the overall performance of different registration approaches. \\
We measure the overlap of the lobes with the Dice coefficient
\begin{equation*}
    DSC = \frac{|X\cap Y|}{|X|+|Y|}
\end{equation*}
where $X$ is the propagated segmentation $b_\M(y)$ and $Y$ is the segmentation of the fixed image $b_\F$. Moreover, we evaluate the average symmetric surface distance
\begin{equation*}
    ASD = \frac{1}{|X_s|+|Y_s|} \left( \sum_{x\in X_S} d(x,Y_S) + \sum_{y\in Y_S} d(y,X_S)\right),
\end{equation*}
where $d$ is the surface distance
\begin{equation*}
    d(x,Y_s) = \min_{y\in Y} d(x,y)
\end{equation*}
where $x$ and $y$ are points on the surface of the propagated segmentation surface $X_s$ and the fixed segmentation surface $Y_s$. Additionally, we calculate the symmetric Hausdorff distance
\begin{equation*}
    HD = \text{max}\{d_H(X_s,Y_S),d_H(Y_s,X_s)\},
\end{equation*}
where
\begin{equation*}
    d_H(X_s,Y_s) = \max_{x_\in X_s} \min_{y\in Y_s} d(x,y).
\end{equation*}

\ \\
We compare our proposed method to the conventional approach of \cite{ruhaak2017estimation} that is currently ranked first in the EMPIRE challenge \citep{murphy2011evaluation} (https://empire10.grand-challenge.org/Home/). This method performs a discrete keypoint detection and matching which are integrated into a dense continuous optimization framework. Additionally to the keypoint penalty, the method uses an NGF distance measure, curvature regularizer, a volume change penalty, and a mask alignment of the lung segmentations. Note that the lung segmentation has to be available for each pair of images to be registered. This is in contrast to our method, which also uses a boundary loss (equation \ref{eq:BoundaryLoss}), but this requires the masks to be only available during training, not during testing.

\subsection{Robustness}
To analyze the robustness of our method, we evaluate the 30\% lowest Dice Scores of all cases. This gives a good overview of the hardest cases and how good our method can register those.

\subsection{Plausibility of the Deformation Field}
Besides accurately and robustly transferring anatomical annotations, medical image registration should also provide plausible deformations and therefore should not generate deformations with foldings. Hence, we evaluate the Jacobian determinant as it is a local measure for volume change and in particular for (local) change of topology. If $\text{det}(\nabla y)>1$ a volume expansion occurred and if $\text{det}(\nabla y)<1$ the volume decreased and for $\text{det}(\nabla y)\leq 0$ there is a folding.\\

\subsection{Applicability}
In a clinical setting, the registration of two scan pairs has to be available quickly in order not to slow down the routine workflow. In other situations such as screenings, the large number of required registration demands efficient deformable image registration methods. In both cases, the runtime of the algorithms is a crucial factor. For the conventional registration method, we measured the time of the registration without the time needed to load and warp the images. For the network, we measure the time of one forward pass through the cascade of networks. Both measurements were run on the same system with an Intel(R) Core(TM) i7-770K CPU and an Nvidia Titan XP GPU.

\subsection{Transferability and Comparison to state-of-the-art}
To show the transferability of our method to other datasets and to compare our method to other registration methods, we apply our trained network as-is to the ten images pairs of the DIRLAB 4DCT. To evaluate the registration accuracy, the target registration error of the landmarks was computed.Moreover, we evaluate the impact of the dataset used to train the registration network. Therefore, we train the widely used Voxelmorph \citep{balakrishnan2019voxelmorph} framework using the COPDGene data. We adapt the public implementation slightly by choosing a higher regularization weight ($\lambda =2$) to obtain smooth deformation fields Furthermore, we applied our trained model on the 30 scan pairs of the EMPIRE10 challenge and submitted the displacement fields to the organizers who performed the evaluation which includes a lung boundaries, fissures, landmarks and singularities (foldings).

\subsection{Ablation Study}
\label{sec:AblationStudy}
In an ablation study, we study the impact of the components of the proposed method. We investigate the influence of the number of levels in the multilevel framework on the accuracy and plausibility of the registration results. For all experiments, the overall number of epochs was 75. Furthermore, we explore the additional penalty terms in our loss function by setting the weighting parameters in the loss function to zero and compare it with the proposed loss function.

\section{Results}
The results of our experiments on the COPDGene dataset are summarized in Table \ref{tab:copdgeneResults}. We performed a one-sided Wilcoxon signed-rank test that show that the improvement to the method of \cite{ruhaak2017estimation} is statistically significant for the Dice score, average surface distance (ASD) and Hausdorff distance(HD) and the runtime on the CPU. In the following subsections, we describe the results of each experiment in more detail.

\begin{table}[t]
\caption{Registration results of \cite{ruhaak2017estimation} and our method on the COPDGene dataset. We performed a one-sited Wilcoxon signed-rank test to test if improvements to the method of \cite{ruhaak2017estimation} are statistically significance. Significance levels are defined as * p $< 0.05$, ** p$< 0.01$ and *** p$<0.001$.} 
    \centering
    \begin{tabular}{c|c|c}
    & \cite{ruhaak2017estimation} & ours \\
    \hline
         Dice & $0.92\pm0.05^{***}$ & $0.95\pm 0.03$\\
         ASD & $1.97\pm 1.24$mm$^{***}$ & $1.72\pm0.89$mm\\
         HD & $27.24\pm13.70$mm$^*$ & $26.84\pm14.27$mm\\
         Dice30 & $0.86\pm 0.03$ $^{***}$& $0.93\pm 0.01$ \\
         Foldings & $0\%$ & $0.06\%$\\
         Runtime CPU &160s $^{***}$ & 32s  \\
         Runtime GPU & - & 0.75s\\
         GPU memory & - & $4$GB
    \end{tabular}
    
    \label{tab:copdgeneResults}
\end{table}

\begin{figure}[H]
\centering
\setlength{\tabcolsep}{0.001\textwidth}
\begin{tabular}{c}
  \includegraphics[width=0.45\textwidth]{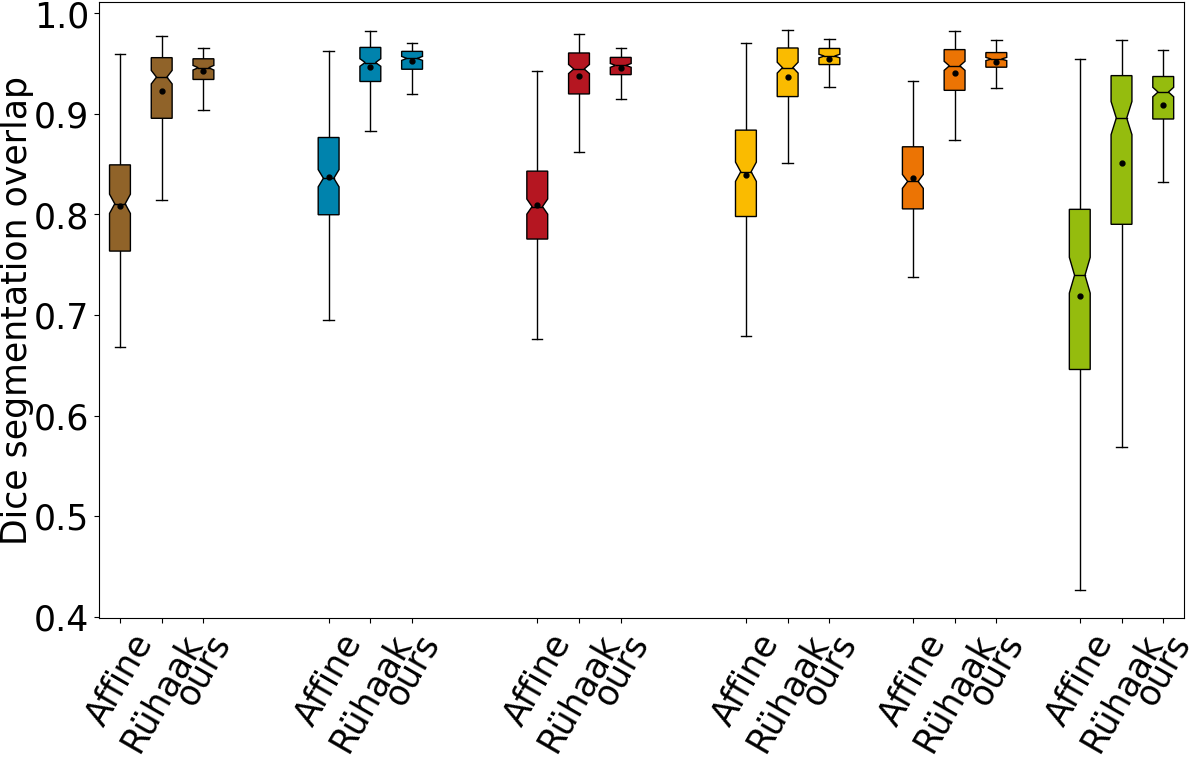}
\\
   \includegraphics[width=0.45\textwidth]{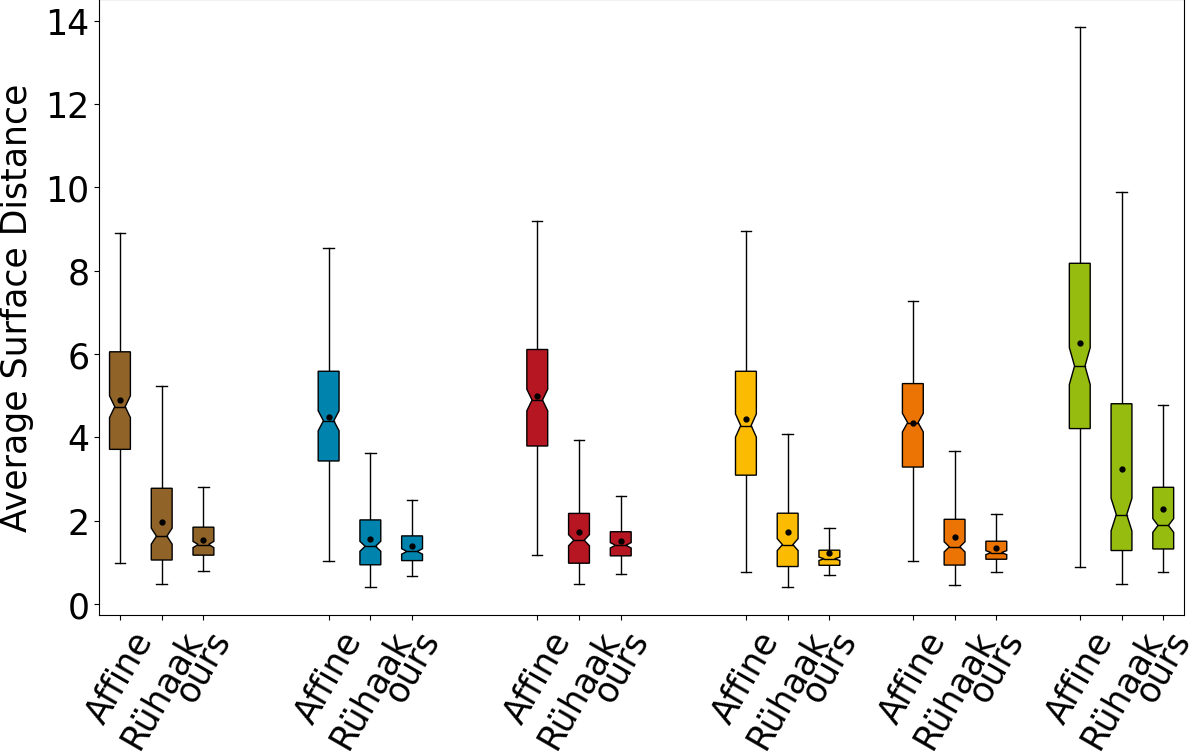}

\\
   \includegraphics[width=0.45\textwidth]{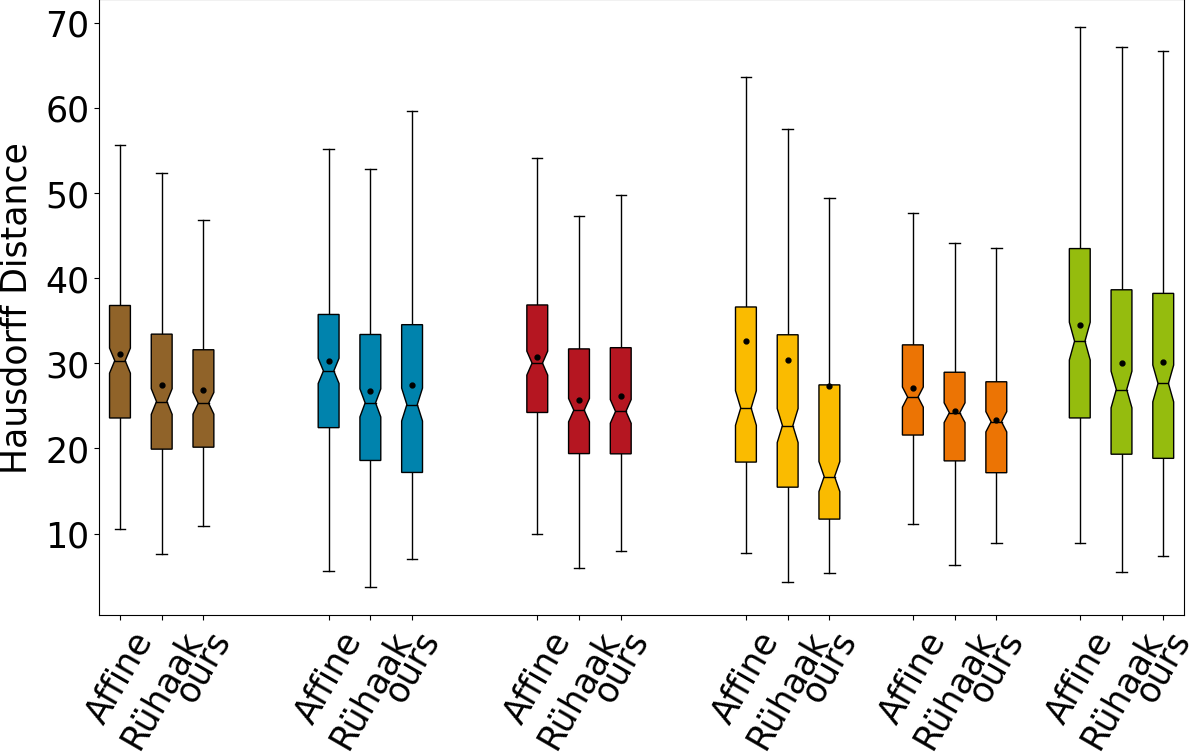}
\end{tabular}
\caption{Comparison of the Dice overlaps, average surface distance and Hausdorff distance for all test images and each anatomical label (\textcolor{Meancolor}{\rule{.2cm}{.2cm}} average of all labels, \textcolor{Lobe1}{\rule{.2cm}{.2cm}} upper left lobe, \textcolor{Lobe2}{\rule{.2cm}{.2cm}} lower left lobe, \textcolor{Lobe3}{\rule{.2cm}{.2cm}} upper right lobe, \textcolor{Lobe4}{\rule{.2cm}{.2cm}} lower right lobe and \textcolor{Lobe5}{\rule{.2cm}{.2cm}} middle right lobe). For each one the distributions of the Dice coefficients after affine pre-alignment, after conventional method of \cite{ruhaak2017estimation} and after our proposed registration are shown.}
\end{figure}
\label{fig:boxplot_results}

\subsection{Accuracy}

%Accurate segmentation of the five pulmonary lobes is important for diagnosis, treatment planning and monitoring for lung diseases such as COPD and fibrosis.
Our proposed method achieves on average significant better Dice Scores than the conventional registration method ($0.95$ vs. $0.92$) with a smaller standard deviation ($0.026$ vs. $0.046$). Also for the symmetric average surface and the Hausdorff distance our method achieves better results ($1.72\pm 0.89$mm vs. $1.97\pm$1.24 and $26.84\pm 14.27$ vs. $27.24\pm13.70$mm, respectively). The distribution of the Dice Scores, of the average surface distance, and of the Hausdorff Distance of both methods are shown in Figures \ref{fig:boxplot_results}. 

\subsection{Robustness}
On the $30\%$ of cases with the lowest Dice Scores, our method achieves an average Dice Score of $0.93\pm 0.01$ within a range of $\left[0.85,0.94 \right]$. Compared to the method of \cite{ruhaak2017estimation} with an average Dice Score of $0.86\pm0.03$ within a range of  $\left[0.78,0.90 \right]$, our method propagates the lobes more robustly.

\subsection{Plausibility}

\begin{figure}[H]
\centering
\setlength{\tabcolsep}{0.001\textwidth}
\begin{tabular}{cc}
  \includegraphics[width=0.22\textwidth]{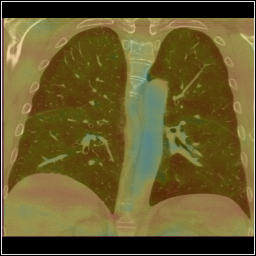}
& \includegraphics[width=0.22\textwidth]{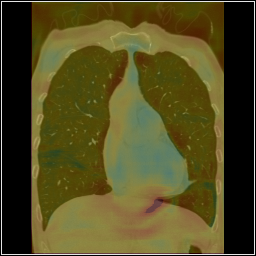}
\\
  \includegraphics[width=0.22\textwidth]{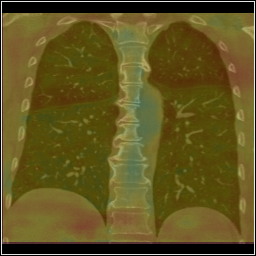}
& \includegraphics[width=0.22\textwidth]{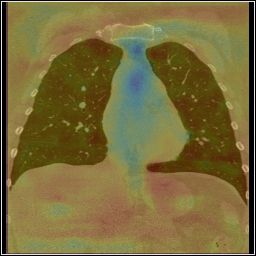} \\
\multicolumn{2}{c}{\includegraphics[width=0.44\textwidth]{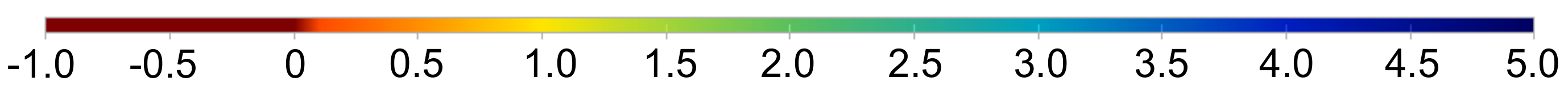}}

\end{tabular}
\caption{Example Jacobian determinant colormaps overlaid on coronal slices of the fixed images. The volume changes are smooth and mostly around 1 (yellow overlay). Due to large motion in the upper right case, some foldings (dark red overlay) occur on the left inferior border.} \label{fig:detJ}
\end{figure}

For our proposed methods, on average fewer than $0.1$\% voxel positions of the deformation field showed a negative Jacobian determinant and therefore a folding. The full elimination of foldings as in the conventional registration methods of \cite{ruhaak2017estimation} is not guaranteed. However, the percentage of foldings is within acceptable ranges. Figure \ref{fig:detJ} shows four exemplary Jacobian determinant colormaps overlaid on the fixed image. The volume changes are smooth and mostly around 1 (yellow overlay). Due to large motion in the upper right case, some foldings (dark red overlay) occur on the left inferior border.

\subsection{Applicability}
The proposed method needs for registration of an image pair on average $0.75$ seconds when executed on a GPU and $32$ seconds on the CPU. In contrast, the conventional method takes on average $160$ seconds executed on a CPU. Moreover, for the execution, only $4$GB of GPU memory are required and therefore our method could also be used on standard computers with less powerful GPUs. The prediction is instantaneous and requires no further manual tuning of parameters. This makes our proposed method very applicable for clinical tasks.

\subsection{Qualitative Results}
To illustrate the registration  results, we show the difference images $\F-\M (y)$ of four exemplary cases in Figure \ref{fig:results}. In all cases, the respiratory motion was successfully recovered and most inner structures are well aligned. The first row shows one example of a more accurately registered image pairs in terms of the average Dice Score (after affine: $0.85$, after: $0.96$) and keypoint distance (after affine: $8.51$mm, after: $0.99$mm). The last row shows the worse case regarding the Dice Score. In this case, the average Dice Score improved from $0.69$ to $0.85$ and the keypoint distance could be reduced from $13.57$mm to $1.9$mm, showing also for the cases with large deformations, our registration methods works robustly. Even with masking the distance measure only to the region inside the lung, the surrounding tissue is mostly well aligned. During training, the model learned to align edges and because no lung mask is given during inference, it also aligns edges outside the lung.

\subsection{Ablation Study}
\begin{figure}[t]
\centering
  \includegraphics[width=0.475\textwidth]{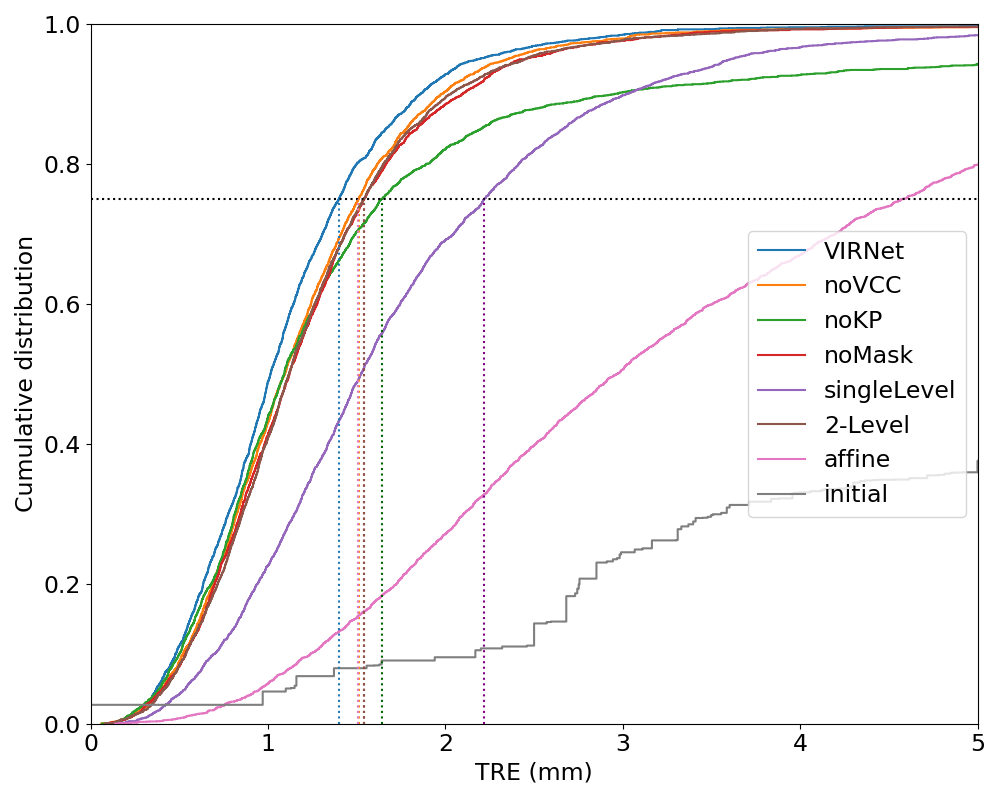}
\caption{Cumulative distribution of target registration error (TRE) in millimeters for all variations of our loss function, after affine registration and initially on all landmark pairs of the DIRLAB 4DCT dataset. In addition, the dotted lines visualize the 75th percentiles of the TRE, which are 1.40mm (our VIRNet), 1.51mm (noVCC), 1.64mm (noKP), 1.54mm (noMask), 2.22mm (singleLevel), 1.54mm (twoLevel), 4.46mm (affine), 12.55mm (initial)} \label{fig:cumLM}
\end{figure}

We provide an ablation study to further verify the efficiency of proposed components of our method. Results of this ablation experiment on the COPDGene data are presented in Table~\ref{tab:ablationStudys}. The multi-level experiment shows that increasing the number of level from $L=1$ to $L=2$ and $L=3$ results in a increasing Dice Score from $0.927$ to $0.939$ to $0.946$, a decreasing TRE fron $3.95$mm to $2.22$mm to $2.00$mm, and decreasing number of foldings from $0.1\%$ to $0.09\%$ to $0.06\%$. The mask alignment loss not only improve the alignment of the pulmonary lobes resulting in a higher Dice Score ($0.93$ vs $0.946$) but also enhance the TRE from $2.16$mm to $2.00$mm. By integrating our volume change loss, the percentage of foldings can be reduced from $0.3\%$ to $0.06\%$. Furthermore, it also improve the TRE from $2.16$mm to $2.00$mm. To further enhance the alignment of smaller structures as vessels and smaller airways, we incorporate keypoint correspondences into the loss function. This decrease the TRE from $4.59$mm to $2.00$mm. However, the percentage of foldings slightly increase from $0\%$ to $0.06\%$. Figure \ref{fig:cumLM} shows a comparison of the target registration errors of the DIRLab 4DCT dataset of all compared settings and after affine registration and the initial errors.

\begin{table*}[t]
    \centering
    \caption{Results of the ablation study. To demonstrate the impact of the each loss function term, each penalty weight was set to zero once while the remaining parameters were fixed to their empirically determined optimal values. The registration performance is evaluated using the Dice score, the target registration error of the keypoint (TRE KP), and the percentage of foldings on the COPDGene dataset. Moreover the target registration error on the DIRLab dataset is compared. We performed a one-sided Wilcoxon signed-rank test to test if improvements to all other settings are statistically significant. We used a Bonferroni correction due to multiples testing. Significance levels are defined as * p $< 0.05$, ** p$< 0.01$ and *** p$<0.001$.} 
    %\scriptsize
    \begin{tabular}{c|c|c|c|c|c|c}
    & no mask align. & no VCC & no keypoint loss & single Level&2-Level &\textbf{proposed} \\
    &  $\beta=0 $ & $\gamma=0$ & $\delta=0$ & $L=1$ &  $L=2 $ &\textbf{settings}  \\
    \hline
         Dice & $0.93\pm0.02$ $^{***}$ & $0.95\pm0.02$ & $0.95\pm0.02$ & $0.93\pm0.02$ $^{***}$& $0.94\pm0.02$ & $0.95\pm0.03$  \\
         TRE KP [mm] & $2.23\pm1.45$ $^{***}$ &$2.16\pm1.34$ $^{***}$ & $4.59\pm2.70$$^{***}$ & $3.95\pm1.98$ $^{***}$ &  $2.22\pm1.43$ $^{***}$ & $2.00\pm1.28$  \\
         Foldings & $0.04\pm0.06$\% & $0.30\pm0.17\%$ $^{***}$ & $0.00\pm0.00\%$ & $0.10\pm0.14\% $$^{**}$ & $0.09\pm0.09\% $ $^{**}$ & $0.06\pm0.03\% $  \\
         TRE 4DCT [mm] & $1.26\pm0.82$ $^{***}$ & $1.22\pm0.84$ $^{***}$ & $1.72\pm2.31$$^{***}$ & $1.76\pm1.11$$^{***}$ & $1.26\pm0.81$$^{***}$ &$1.14\pm0.76$ \\
    \end{tabular}
    \label{tab:ablationStudys}
\end{table*}

\subsection{Transferability and Comparison to state-of-the-art}
In Table \ref{tab:dirlab_convDL}, quantitative results on the DIRLAB 4DCT dataset of deep-learning-based and conventional registration methods are reported. On average the target registration error (TRE) of our method was $1.14\pm0.76$mm and is therefore better as the currently best deep-learning-based method of \cite{fu2020lungregnet}. In cases 6, 8, and 10 which have a large initial landmark error, our method achieves much better registration results. Training Voxelmorph on the large COPDGene dataset results in a lower TRE than when trained by leave-one-out on the DIRLAB dataset ($1.71$mm vs $3.65$mm). The best conventional registration method of \cite{ruhaak2017estimation} has still a lower TRE, however, it needs about 5 minutes to compute the deformation field, whereas our method only needs less than a second. A detailed evaluation of all ten cases for different deep-learning-based registration methods is given in Table \ref{tab:dirlab_cases}. On the EMPIRE10 challenge data, our method achieves a target registration error of $1.01$mm on all cases and a TRE of $0.91$mm if ovine data is excluded. A summary of the results is shown in Table \ref{tab:empire10} and a more detailed evaluation on the challenge website \footnote{\url{https://empire10.grand-challenge.org/mevis_virnet/}}.

\begin{table}[t]
    \centering
    \caption{Results of the EMPIRE10 challenge for the method of \cite{ruhaak2017estimation} and our proposed method. The average score over all 30 cases for the lung boundaries, fissure alignment, landmark error and singularities is shown. Detailed results can be found on the challenge website.} \label{tab:empire10}
    %\scriptsize
    \begin{tabular}{c|c|c|c|c|c}
    
    & Lung B. & Fissures & Landmarks & Singularities \\
    \hline
    Rühaak & 0.00 & 0.09  & 0.63 & 0.00 \\
    ours &0.07 & 0.09 & 1.01 & 0.02
    \end{tabular}
\end{table}
\begin{center}
\begin{table*}[t]

\caption{Target registration error values for different conventional and deep learning-based methods on DIRLAB 4D-CT dataset. All results were extracted from the original papers, besides Voxelmorph* which was reported in \cite{hansen2020tackling} and Voxelmorph** which we trained on the COPDGene data. Since the runtime was not measured with the same hardware, it cannot directly be compared. However, it gives an impression of the speed.}
 \centering
\begin{tabular}{ c | c | c |  c | c }

& Method & mean TRE (mm) & Foldings & Runtime \\
\hline
& initial & $8.46\pm 6.58$ & - & - \\
\hline
&  \cite{schmidt2011landmark} & $1.38\pm0.87$ & - & 83min \\
&  \cite{heinrich2012deeds} & $1.6\pm1.7$ &$ 0\%$ & 20min \\
\textbf{Conventional} & \cite{heinrich2013mrf} & $1.43\pm 1.3$ & - & $7.97$min \\
&  \cite{berendsen2014registration} & $1.36\pm0.99$ & 0\% & - \\
& \cite{ruhaak2017estimation}  & $0.94 \pm 1.06$ & 0\% & 5min \\
\hline
&  \cite{sentker2018gdl} & $2.5\pm 1.16$ & - & few seconds \\
& \cite{balakrishnan2019voxelmorph}$^*$ & $3.65\pm2.47$ & - & - \\
& \cite{balakrishnan2019voxelmorph}$^{**}$ & $1.71\pm2.86$ & - & - \\
&  \cite{deVos2019deep} & $2.64\pm 4.32$  & - & $0.63$s \\
\textbf{Deep Learning}&   \cite{eppenhof2019progressively} & $2.43\pm 1.81$ & $0.42\%$ & $0.56$s \\
&  \cite{hering2019mlvirnet} & $2.19\pm1.62$ & - & - \\
&  \cite{hansen2020tackling}& $1.97\pm 1.42$ & - & - \\
&  \cite{jiang2020multi}& $1.66\pm 1.44$ & $<0.1\%$ & 1.4s \\
& \cite{fu2020lungregnet}& $1.59\pm 1.58$ & - & 1min \\
& \cite{hansen2021graphregnet} & $1.39\pm 1.29$ &$0.02\%$ & $2$s\\ 
& \textbf{ours} & $1.14\pm 0.76$ & $<0.0005\% $& $0.75$s \\
\end{tabular}
\label{tab:dirlab_convDL}
\end{table*}
\end{center}

\begin{table*}[t]
%\scriptsize
\caption{Mean (standard deviation) of the registration error in mm determined on DIR-Lab 4D-CT data for several deep-learning-based registration methods: \cite{eppenhof2019progressively},  \cite{deVos2019deep}, \cite{hering2019mlvirnet}, \cite{fu2020lungregnet}, \cite{hansen2021graphregnet} and  \cite{balakrishnan2019voxelmorph} (trained on the COPDGene data). Asterisk indicates that the FineNet was performed twice for this case and method. }
\begin{tabular*}{\linewidth}{ c|c|c|c|c|c|c|c|c}
%\toprule
\hline
Scan & Initial &  Eppenhof & DLIR  & mlVIRNet& LungRegNet & GraphRegNet &Voxelmorph   & \textbf{ours}\\
\hline%\midrule
4DCT 01 & $3.89\pm2.78$ & $2.18\pm1.05$  & $1.27\pm1.16$  & $1.33\pm0.73$ & $0.98\pm 0.54 $& $0.86$ &$1.03\pm 1.01$ & $0.99\pm 0.47$\\
4DCT 02 & $4.34\pm3.90$ &$2.06\pm0.96$ & $1.20\pm1.12$  &$1.33\pm0.69$ & $0.98\pm 0.52$&$0.90$ &$1.09\pm 1.87$  & $0.98\pm0.46$\\
4DCT 03 & $6.94\pm4.05$ &$2.11\pm1.04$ &$1.48\pm1.26$  &$1.48\pm0.94$ & $1.14\pm 0.64$ &$1.06$ &$1.40\pm 2.04$  & $1.11\pm0.61$\\
4DCT 04&$ 9.83\pm4.85$ & $3.13\pm1.60$ & $2.09\pm1.93$ &$1.85\pm1.37$ & $1.39\pm 0.99$ &$1.45$ &$1.69\pm 2.60$  &  $1.37\pm1.03$\\
4DCT 05 & $7.48\pm5.50$& $2.92\pm1.70$& $1.95\pm2.10$&$1.84\pm1.39$  & $1.43\pm 1.31$ &$1.60$ &$1.63\pm 2.44$ & $1.32\pm1.36$\\
4DCT 06 & $10.89\pm6.96$&$4.20\pm2.00$ & $5.16\pm7.09$&  $3.57\pm2.15$ & $2.26\pm 2.93^*$&$1.59$ &$1.60\pm 2.58$  & $1.15\pm1.12$\\
4DCT 07& $11.03\pm7.42$&$4.12\pm2.97$ & $3.05\pm3.01$& $2.61\pm1.63$ & $1.42\pm 1.16^*$&$1.74$ &$1.93\pm 2.8$  & $1.05\pm0.81$ \\
4DCT 08 &$14.99\pm9.00$& $9.43\pm6.28$& $6.48\pm5.37$ &$2.62\pm1.52$ & $3.13\pm 3.37^*$ &$1.46$ &$3.16\pm 4.69$  & $1.22\pm1.44$\\
4DCT 09 &$7.92\pm3.97 $& $3.82\pm1.69$& $2.10\pm1.66$&$2.70\pm1.46 $& $1.27\pm 0.94$& $1.58$ &$1.95\pm 2.37$  & $1.11\pm0.66$\\
4DCT 10 & $7.30\pm6.34$&$2.87\pm1.96$ & $2.09\pm2.24$ &$2.63\pm1.93 $ & $1.93\pm 3.06$ &$1.71$&$1.66\pm 2.87$  & $1.05\pm0.72$\\
\hline
Mean & $8.46\pm6.58$ & $3.68\pm3.32$& $2.64\pm4.32$& $2.19\pm1.62$ & $1.59\pm 1.58$ &$1.39\pm 1.29$ &$1.71\pm 2.86$ &  $1.14\pm0.76$\\

\end{tabular*}
\normalsize
\label{tab:dirlab_cases}
\end{table*}

\begin{figure*}[t]
\centering
\setlength{\tabcolsep}{0.001\textwidth}
\begin{tabular}{ccccc}
  \includegraphics[width=0.19\textwidth]{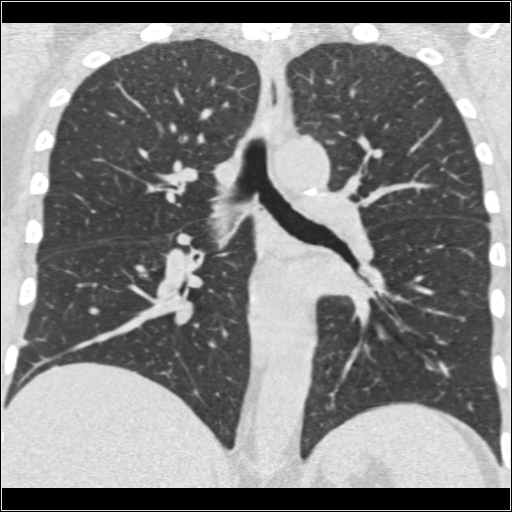}
& \includegraphics[width=0.19\textwidth]{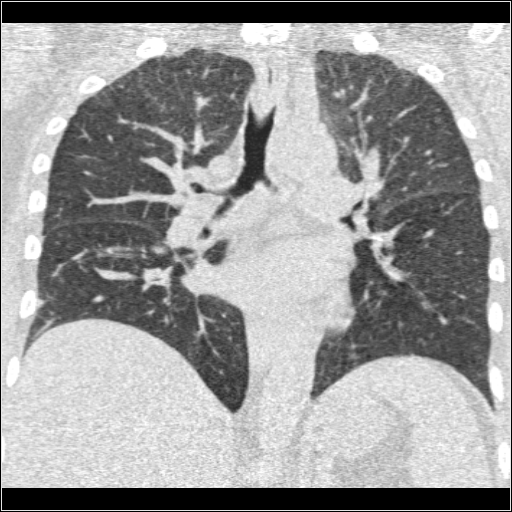}
& \includegraphics[width=0.19\textwidth]{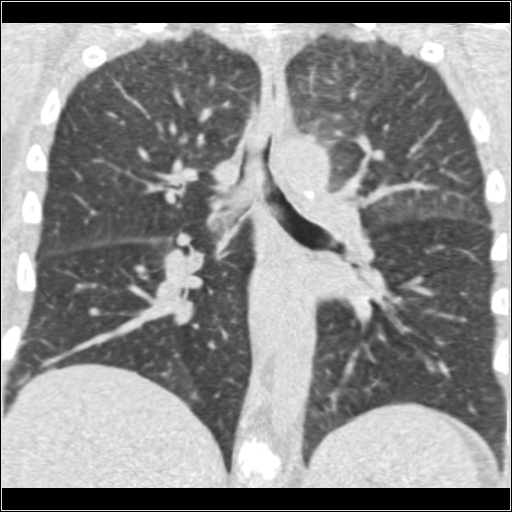}
& \includegraphics[width=0.19\textwidth]{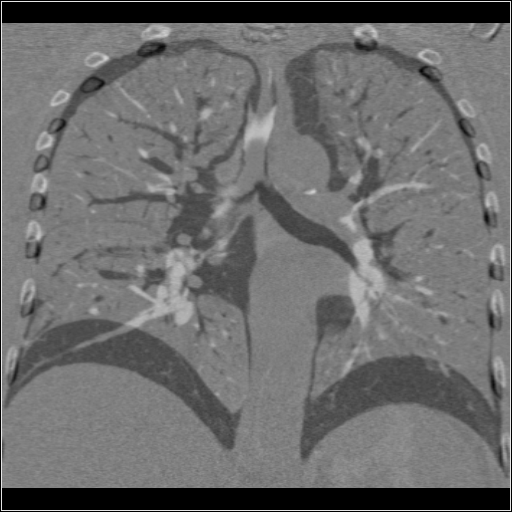}
& \includegraphics[width=0.19\textwidth]{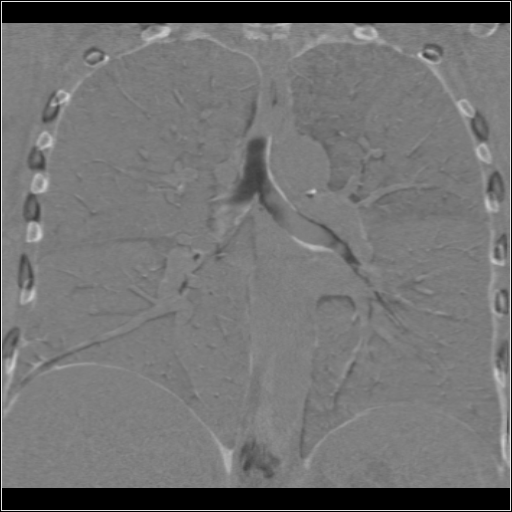}
\\
  \includegraphics[width=0.19\textwidth]{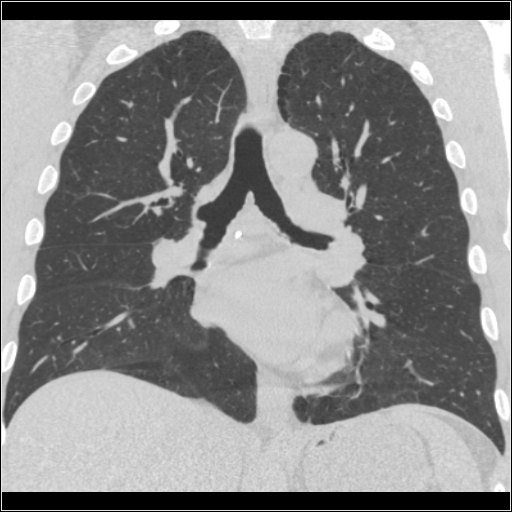}
& \includegraphics[width=0.19\textwidth]{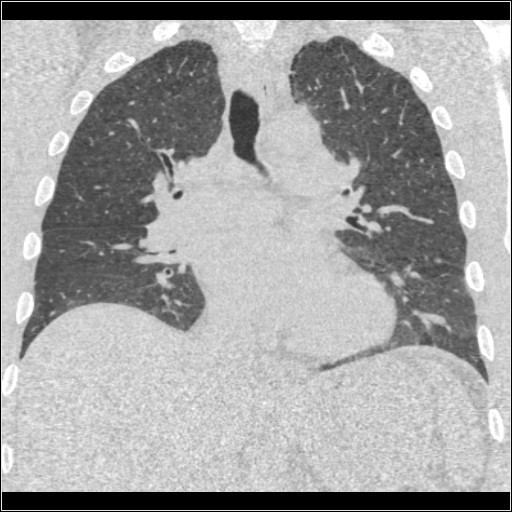}
& \includegraphics[width=0.19\textwidth]{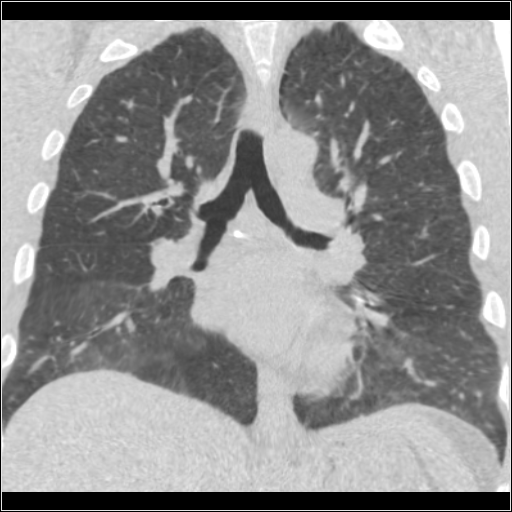}
& \includegraphics[width=0.19\textwidth]{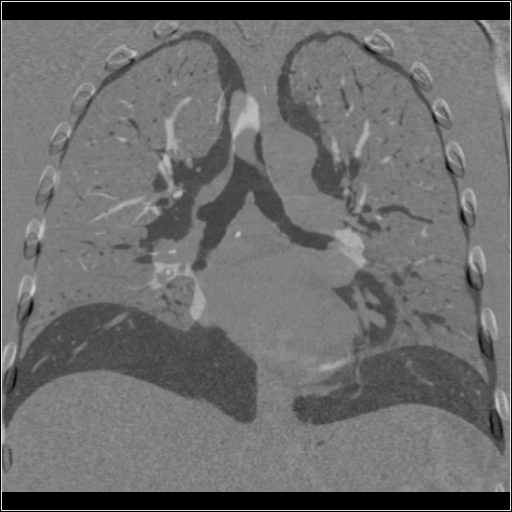}
& \includegraphics[width=0.19\textwidth]{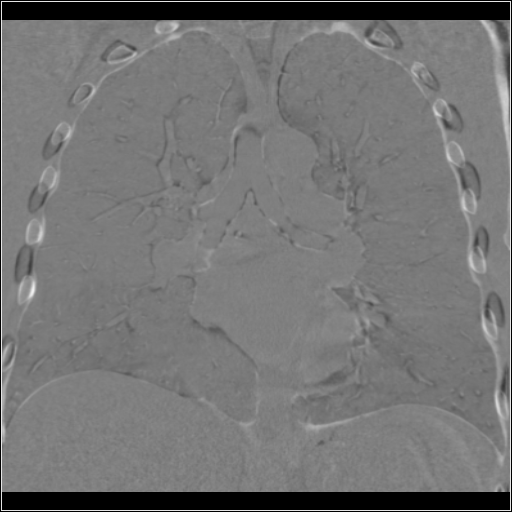}
\\
  \includegraphics[width=0.19\textwidth]{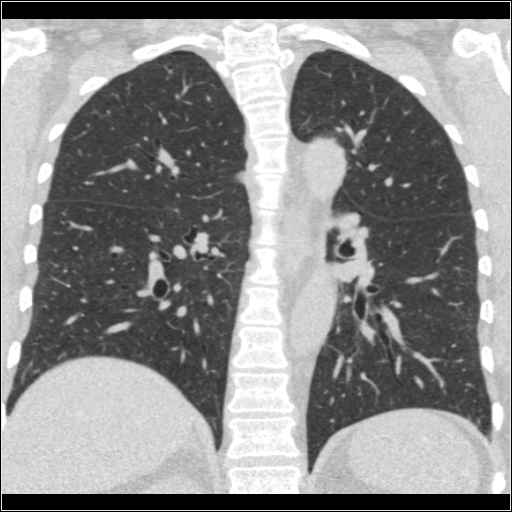}
& \includegraphics[width=0.19\textwidth]{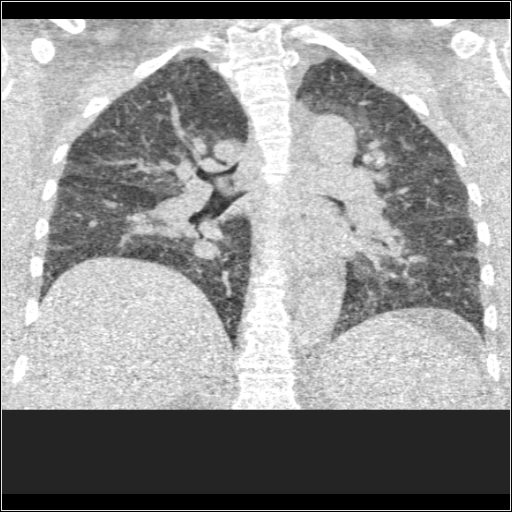}
& \includegraphics[width=0.19\textwidth]{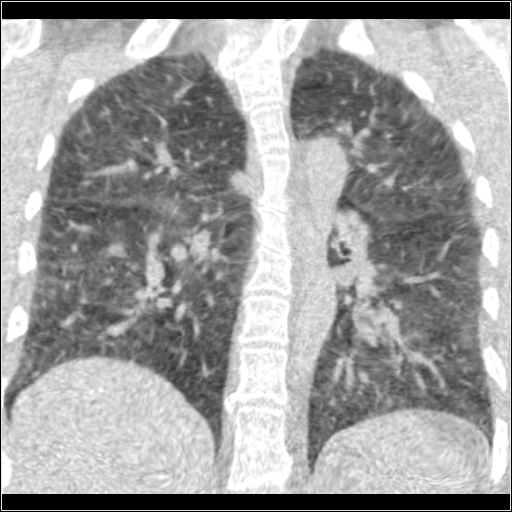}
& \includegraphics[width=0.19\textwidth]{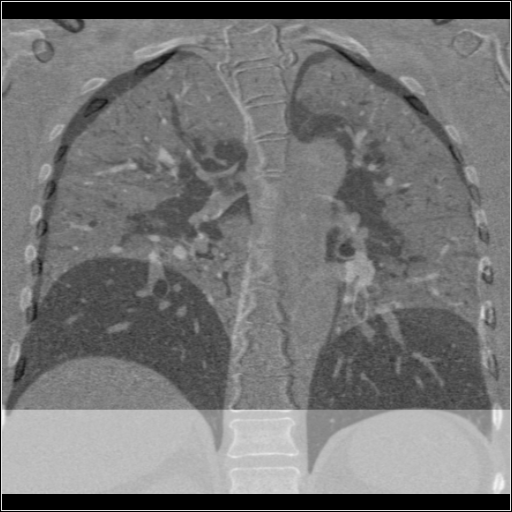}
& \includegraphics[width=0.19\textwidth]{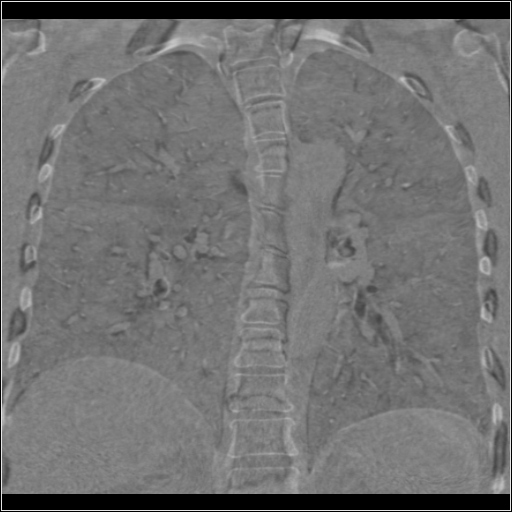}
\\
  \includegraphics[width=0.19\textwidth]{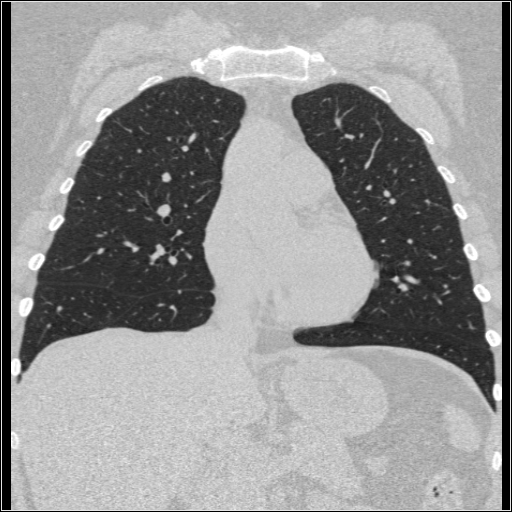}
& \includegraphics[width=0.19\textwidth]{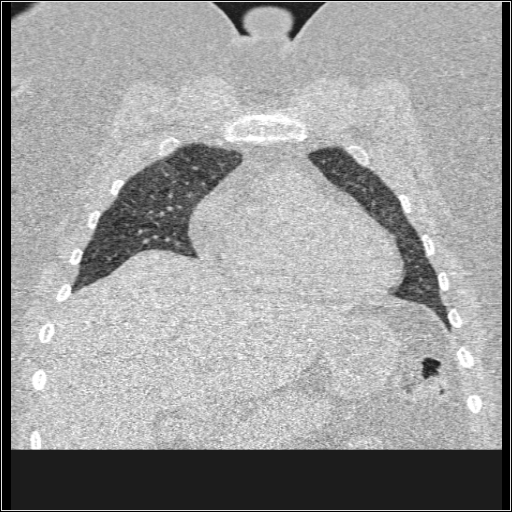}
& \includegraphics[width=0.19\textwidth]{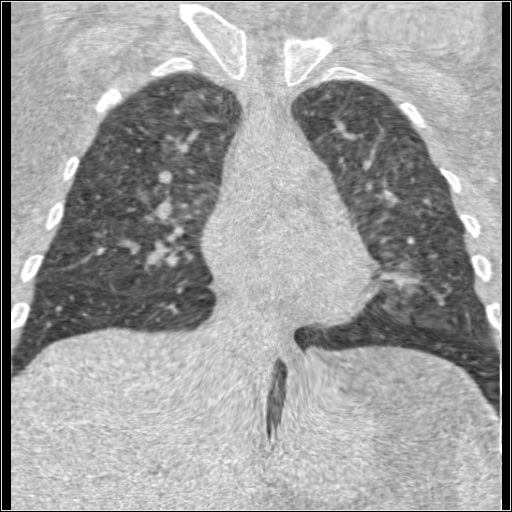}
& \includegraphics[width=0.19\textwidth]{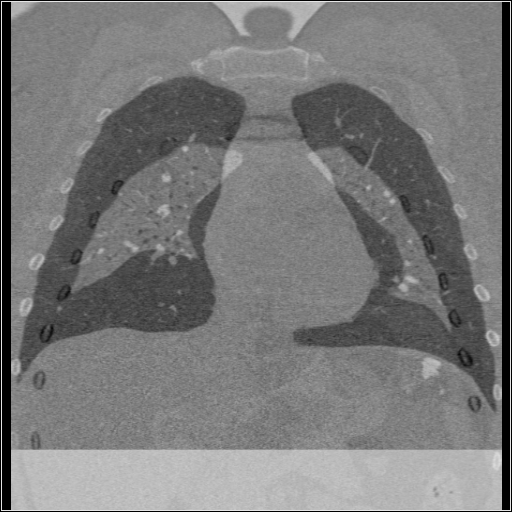}
& \includegraphics[width=0.19\textwidth]{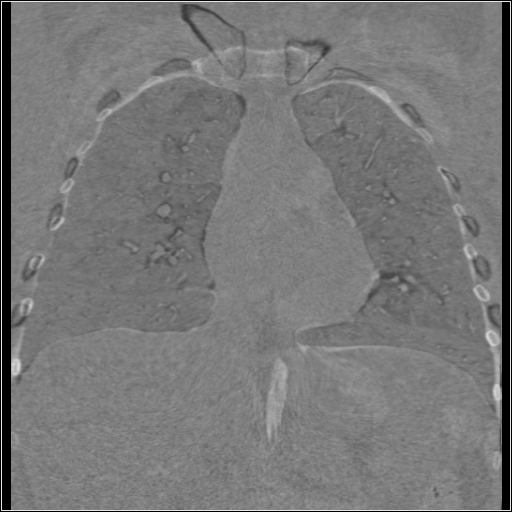}
\\
  {Fixed $\F$}
& {Moving $\M$}
& {Warped $\M(y)$}
& {$\F$-$\M$}
& {$\F$-$\M(y)$}
\end{tabular}
\caption{Example coronal slices extracted from four exemplary cases. Input images $\F$ and $\M$, the warped moving image $\M(y)$, the difference image $\F-\M$ (fourth column) and the difference image $\F-\M (y)$ after registration with the proposed method (fifth column). In all cases the respiratory motion was successfully recovered and most inner structures are well aligned. Due to altered density of lung tissue during breathing, intensity changes occur and therefore higher values in the difference images are reached without registration errors. } \label{fig:results}
\end{figure*}

\section{Discussion}
This paper presents a coarse-to-fine multilevel framework for deep-learning-based image registration that can compensate for and handle large deformations using computing deformation fields on different scales. Our method shares many elements with the conventional registration method of \cite{ruhaak2017estimation}. We have identified key strategies of this method and successfully developed a deep-learning counterpart. The advantage of our deep learning approach is that the expensive annotation and detection of the lobe masks and keypoints is only necessary as training data. This important knowledge is then embedded in our model and therefore the inference is cheap and fast. \\

We employ a Gaussian-pyramid-based multilevel framework that can solve the image registration optimization in a coarse-to-fine fashion. To prevent foldings of the deformation field and restrict the determinant of the Jacobian to physiologically meaningful values, we combine the curvature regularizer with a volume change penalty in the loss function. Furthermore, we also integrate weak keypoint correspondences into the loss function to focus more on the alignment of smaller structures. The keypoints are computed automatically and can be considered as noisy labels with residual errors of $1-2$mm. However, we showed that the use of these noisy labels is nevertheless advantageous and leads to a better alignment of vessels and smaller airways and therefore also results in a better target registration error on the DIRLab dataset.

We validated our framework on the challenging task of large motion inspiration-to-expiration registration using image data from the multi-center COPDGene study. To assess the accuracy of our network, we performed an extensive evaluation of 200 pulmonary CT scan pairs from the large-scale COPDGene study and demonstrated that our method can perform accurate registration between two affine pre-aligned images. Especially for the task of lobe propagation, we could show that our method performs better than conventional approaches. It achieves higher Dice scores and lower surface and Hausdorff distances ($0.95$, $1.72$~mm, and $26.8$~mm) compared to conventional registration ($0.92$, $1.97$~mm, and $27.2$~mm, respectively). This better performance can be explained by the use of the mask-alignment loss. As demonstrated in previous studies (e.g. \cite{balakrishnan2019voxelmorph,hering2019memory}), the combination of the complementary strength of global semantic information (weakly-supervised learning with segmentation labels) and local distance metrics improves the registration performance during inference. In contrast to conventional registration methods, such additional information only needs to be available in the training dataset.

Furthermore, we have evaluated the proposed method using the DIRLab and EMPIRE10 dataset and showed that we achieve excellent TRE of 1.14~mm and 1.01~mm, respectively. Note that our network was not trained on those datasets.  This is strong evidence that our network can generalize well. Although previous works (e.g. \cite{deVos2019deep,eppenhof2019progressively, jiang2020multi,sentker2018gdl}) contribute much to improving the registration accuracy, there is still a misalignment of smaller structures, which leads to a high TRE. To focus more on the alignment of vessels, \cite{fu2020lungregnet} introduced a preprocessing step to enhance the vessels in the input images by segmenting vascular structures and increasing the intensity value inside the vessel mask. In their paper, they demonstrated the efficiency of this preprocessing step. Since this step is performed on the input images, it is also required during application. To avoid this problem and thus not increase the execution time, we integrate additional information on smaller structures using the keypoint loss. The advantage of this procedure is that the keypoints, as well as the masks of the boundary loss, are only needed during training. Nevertheless, the best conventional registration methods still achieve lower TRE than our method. One reason for this might be that convectional registration methods mostly work on the original image data. In contrast, for the deep-learning approaches, the input images have to be downsampled due to memory restrictions on the GPU. Especially for smaller structures and small errors (we are speaking about a TRE difference of 0.2-0.4mm), it is easily imaginable that this resolution is not high enough. Moreover, the training data used also influence the performance. Our network was trained on inspiration-expiration scan pairs from humans. In the EMPIRE10 dataset, a variety of lung registration tasks including ovine lung registration has to be performed. Although our method does not register the ovine data perfectly, we achieve a TRE of 1.69~mm on the ovine data which shows that our method is capable of generalizing well. We would assume that with a wider variety in training data, the performance of deep-learning-based registration methods can further improve. We showed this effect when training the Voxelmorph network. By using the larger COPDGene dataset to train the Voxelmorph network, the target registration error on the DIRLab dataset improved from $3.65$~mm to $1.71$~mm compared to a leave-one-out training on the DIRLab dataset. This illustrates the large impact of the training dataset. Since Voxelmorph and our framework are very similar, this experiment also shows that the addition of more loss functions and a multilevel approach is beneficial.

Besides accurately transferring anatomical annotations, medical image registration should also provide plausible transformations and therefore should not generate deformations with foldings. In conventional registration methods, this is achieved by using a regularizer in the cost function. Recently deep-learning-based methods like \cite{eppenhof2019progressively} and \cite{jiang2020multi} also integrated a regularizer into their loss functions to enforce a smooth deformation field resulting in an acceptable amount of foldings ($0.42\%$ and $~0.1\%$ of foldings). In our work, we additionally use a volume change control which penalizes occurring foldings more severely than the regularizer does, resulting in on average fewer than $0.1\%$ and $0.0005\%$ voxel positions in the deformation field with folding on the COPDGene and DIRLab dataset, respectively. Without the volume change control penalty, the deformation fields produced by our method show on average $0.30\%$ of voxel positions with foldings, which is comparable to the values of other deep learning registration methods. This shows that the addition of the volume change control mitigated the occurrence of foldings. The higher number of foldings on the COPDGene dataset can be explained by the noise difference between the expiration and inspiration scan due to different doses during acquisition (see Fig \ref{fig:results} for some example images). The full elimination of foldings as in some conventional registration methods is not guaranteed. Another way to reduce the number of foldings was presented in the works of \cite{dalca2018diffeomorphic}, \cite{krebs2019unsupervised}, and \cite{qiu2021learning} who are using the scaling and squaring algorithm \citep{arsigny2006ssalgorithm} to integrate the predicted stationary velocity field. With a sufficient number of integration steps, these methods should theoretically guarantee diffeomorphic transformation. However, in the presented works they reported "nearly no non-negative Jacobian voxels" \citep{dalca2018diffeomorphic} and $0.023\%$ to $0.151\%$ of voxels with a negative Jacobian determinant \citep{qiu2021learning}. As discussed in \citep{qiu2021learning}, this has two major factors. First, the velocity field could be not sufficiently smooth. This can be solved by increasing the regularization weight. However, this often yields a drop in the registration accuracy. Secondly, the number of chosen integration steps was too small. Increasing this can reduce the number of foldings which occur but increase the computational cost as well. In summary, the scaling and squaring approach and the  volume-change-control penalty presented achieve similar results preventing foldings. Besides, our approach regulates volume changes.

In our experiments, we focused on the complex task of CT lung registration, as the registration results can be evaluated more accurately than only with an overlap of a larger structure. However, our method could also be trained for a different task or on a different modality. Except for keypoint detection, no component is lung-specific and the keypoint loss can be used with landmarks in different organs. \\
In future studies, we will investigate the impact of instance optimization to fine-tune the deformation field for those image pairs for which the registration result is not yet satisfactory.

\section{Conclusion}
This paper presents a deep-learning-based registration approach for deformable image registration, targeting in particular the challenging task of lung registration. We introduce a keypoint matching term and a volume change penalty to increase the alignment of smaller structures and to prevent foldings and restrict the deformation field to physiologically meaningful values. Our multi-level registration framework equipped with these components achieves state-of-the-art registration accuracy on the COPDGene and DIRLab datasets with a very short execution time.

\section*{Declaration of Competing Interest}
There’s no financial/personal interest or belief that could affect the objectivity of the submitted research results. No conflict of interests exist.

\section*{Acknowledgements}
The authors are deeply grateful to Keelin Murphy, Edward Castillo and Richard Castillo for providing evaluation benchmarks. \\
This work was supported by the German Academic Scholarship Foundation. We  gratefully  acknowledge  the  COPDGene  Study  for providing the data used.  COPDGene is funded by Award Number U01 HL089897 and Award Number U01 HL089856 from the National Heart, Lung, and Blood Institute. COPDGene is also supported by the COPD Foundation through contributions made to an Industry Advisory Board comprised of AstraZeneca, Boehringer-Ingelheim, Genentech, GlaxoSmithKline, Novartis, Pfizer, Siemens, and Sunovion. We gratefully acknowledge the support of the NVIDIA Corporation with their GPU donations for this research.
%%Harvard
\bibliographystyle{model2-names.bst}\biboptions{authoryear}
\bibliography{hering.bbl}

\end{document}